\documentclass[times,twocolumn,final]{elsarticle}

\usepackage{algorithm2e}
\usepackage{amsmath}
\usepackage{amssymb}
\usepackage{medima}
\usepackage{framed,multirow}
\usepackage{color}
\usepackage{latexsym}
\usepackage{comment}
\usepackage{url}
\usepackage{mathtools}
\usepackage{xcolor}
\usepackage{multirow}
\usepackage{booktabs}
\usepackage{hyperref}
\DeclareUnicodeCharacter{2009}{\,} 
\definecolor{newcolor}{rgb}{.8,.349,.1}
\usepackage{tabularx}
\journal{Medical Image Analysis}

\begin{document}

\verso{Tong \textit{et~al.}}

\begin{frontmatter}

\title{CLANet: A Comprehensive Framework for Cross-Batch Cell Line Identification Using Brightfield Images}%

\author[1,2]{Lei Tong}
\author[2]{Adam Corrigan}
\author[3]{ Navin Rathna Kumar}

\author[3]{Kerry Hallbrook}
\author[4]{Jonathan Orme}
\author[2]{Yinhai Wang\corref{cor1}}

\ead{yinhai.wang@astrazeneca.com}
\author[1]{Huiyu Zhou\corref{cor1}}
\ead{hz143@leicester.ac.uk}
\cortext[cor1]{Corresponding author.
}

\address[1]{School of Computing and Mathematical Sciences, University of Leicester, Leicester, UK}
\address[2]{Data Sciences and Quantitative Biology, Discovery Sciences, AstraZeneca R$\&$D, Cambridge, UK}
\address[3]{UK Cell Culture and Banking, Discovery Sciences, AstraZeneca R$\&$D, Alderley Park, UK}
\address[4]{UK Cell Culture and Banking, Discovery Sciences, AstraZeneca R$\&$D, Cambridge, UK}


\begin{abstract}
Cell line authentication plays a crucial role in the biomedical field, ensuring researchers work with accurately identified cells. Supervised deep learning has made remarkable strides in cell line identification by studying cell morphological features through cell imaging. However, batch effects, a significant issue stemming from the different times at which data is generated, lead to substantial shifts in the underlying data distribution, thus complicating reliable differentiation between cell lines from distinct batch cultures. To address this challenge, we introduce CLANet, a pioneering framework for cross-batch cell line identification using brightfield images, specifically designed to tackle three distinct batch effects. We propose a cell cluster-level selection method to efficiently capture cell density variations, and a self-supervised learning strategy to manage image quality variations, thus producing reliable patch representations. Additionally, we adopt multiple instance learning(MIL) for effective aggregation of instance-level features for cell line identification. Our innovative time-series segment sampling module further enhances MIL's feature-learning capabilities, mitigating biases from varying incubation times across batches. We validate CLANet using data from 32 cell lines across 93 experimental batches from the AstraZeneca Global Cell Bank. Our results show that CLANet outperforms related approaches (e.g. domain adaptation, MIL), demonstrating its effectiveness in addressing batch effects in cell line identification.

\end{abstract}

\begin{keyword}
\KWD Cell line authentication\sep Brightfield image analysis\sep Batch effect\sep Multiple instance learning
\end{keyword}

\end{frontmatter}


\section{Introduction}
\label{sec1}
Cell line authentication (CLA) is a crucial process for verifying the identity of cell lines employed in scientific research. This verification is necessary as cell lines can be mislabelled, contaminated, or undergo changes over time, potentially affecting the reliability and reproducibility of research findings (\cite{ioannidis2005most,boonstra2010verification}). At present, short tandem repeat (STR) profiling serves as the gold standard for authentication, utilized to confirm a cell line's identity. However, this method has its limitations (\cite{masters2001short,freedman2015reproducibility}). Long-term culture, subcloning, and selection can instigate microsatellite instability and loss of heterozygosity, particularly in cancer cell lines, possibly leading to genetic drift that traditional STR methods struggle to detect (\cite{parson2005cancer}). Further, isogenic cell lines pose an authentication challenge using this method (\cite{reid2013authentication}). Also, owing to time and cost constraints, the standard practice involves testing the cells once they are fully expanded and frozen. Nevertheless, if the sample fails STR profiling, this process results in wasted time and resources. 

\begin{figure*}
    \centering
    \includegraphics[width = 1\textwidth]{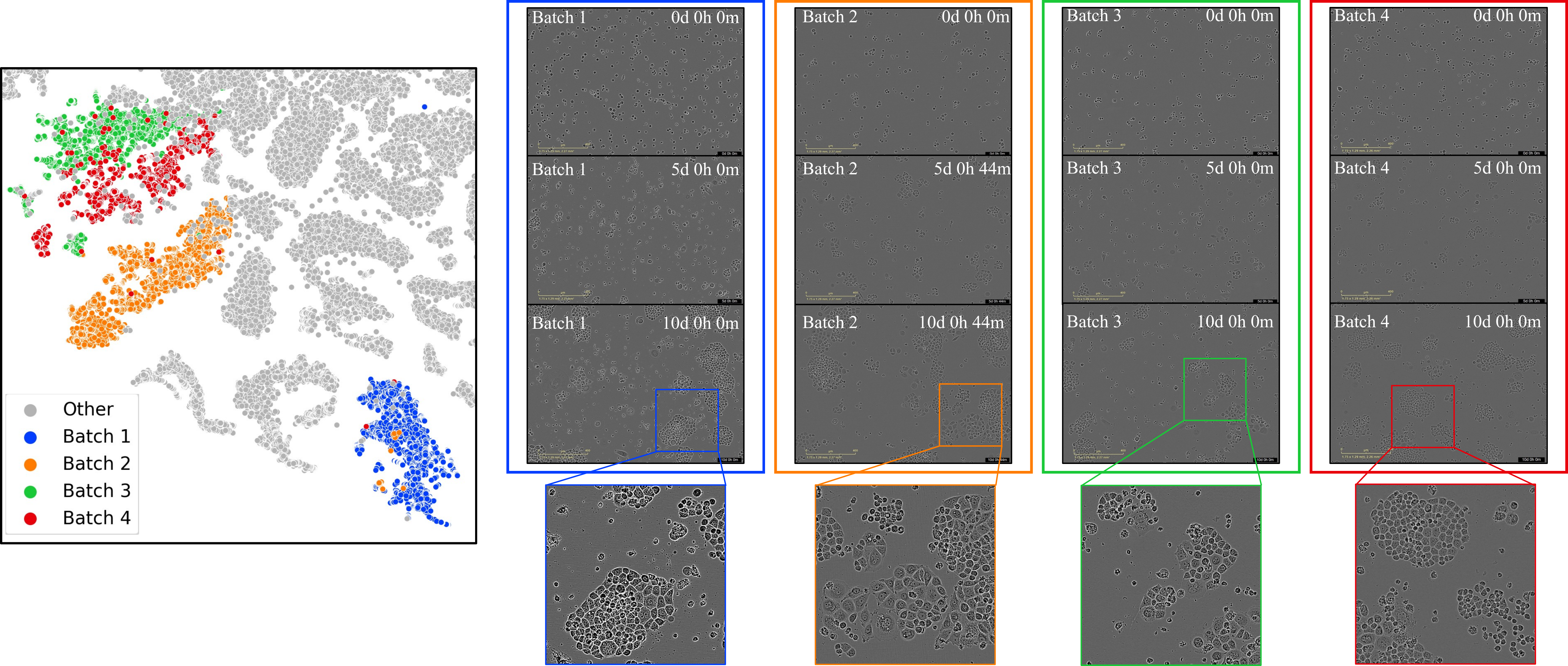}
    \caption{Illustration of batch effects in cell line image data. The t-SNE plot depicts cell images as points, with colorful points representing four distinct batches from CAMA1 cell lines and grey points representing data from other cell lines. In the subsequent columns, 3 example images are provided for each batch. The batch ID and the time point of incubation are labeled with white text in each image.}
    \label{fig:batch_effect} 
\end{figure*}

Advancements in machine learning (ML) techniques have significantly contributed to the development of image-based cell profiling (ICP) as a promising approach for rapid, cost-effective analysis, while simultaneously providing the potential to detect changes in cell morphology indicative of undesirable attributes such as genetic drift or cellular senescence. The typical workflow of ICP involves capturing cellular images using microscopy techniques, followed by the extraction of morphological features from the images, and finally, conducting analyses with computational models tailored to the specific target application (\cite{caicedo2016applications}). Existing works have achieved great success in detecting genetic perturbations based on fluorescent images or multi-channel images derived from cell painting assays (\cite{caicedo2017data,chandrasekaran2021image,cross2022self}). However, the phototoxicity of fluorescence can adversely affect cell growth, rendering it challenging to acquire cellular images over extended periods. Brightfield imaging offers numerous advantages over fluorescent labeling, including simplified setup with only basic equipment required and the ability to track cell genealogies in long-term time-series experiments. In particular, brightfield images yield the possibility to measure more descriptive features such as texture and shape simultaneously (\cite{buggenthin2013automatic}). Researchers have utilized brightfield imaging to identify four cancer cell lines (COLO 704, EFO-21, EFO-27, UKF-NB-3) and their sublines adapted to the anti-cancer drugs (\cite{mzurikwao2020towards}). Recent work reported that it is feasible to  
identify cell lines and predict incubation durations based on brightfield images (\cite{tong2022automated}). Nevertheless, despite its low cost and ease of use, brightfield imaging has limitations in resolving fine details of cells and their structures, particularly in thick specimens, due to constraints in contrast and resolution. Furthermore, batch effects can introduce significant challenges for image-level analysis of cells.

Batch effects\footnote{In cell culture, a batch of data refers to a specific cell line that is incubated in one flask for several days.} are inevitable in biomedical studies, as data are often generated at different times, leading to confounding factors that can obscure biological variations (\cite{li2020deep}). These effects can lead to substantial changes in the underlying data distribution. As illustrated in Fig. \ref{fig:batch_effect},  we showcase t-SNE embeddings of images from 4 distinct batches of the CAMA1 cell line. Each of the images from these batches is initially transformed into 1536-dimensional features using a Vision Transformer (ViT) backbone, with only the data from batch 1 being used for training the model on a multi-class image classification task. Then, the extracted features of the 4 batches of data are mapped to the t-SNE plot. Despite the 4 batches of data all belonging to the same class, the data from the same batch tend to cluster together, distinct from data originating from other batches. It also clearly demonstrates both large intra-class variation between the four batches of data and small inter-class variation between these batches and other cell lines. Failure to address batch effects can result in substantial prediction errors in downstream analyses. Prior studies proposed to reduce the influence of batch effects by employing domain adaptation techniques and incorporating multiple batches during the training of prediction models (\cite{kothari2013removing,cross2022self,sypetkowski2023rxrx1}). In addition, considering that cell images collected from the same batch share the same class labels, aggregating instance-level features to generate batch-level predictions may further decrease the likelihood of prediction errors. 

In this paper, we propose a comprehensive framework, namely CLANet, for cross-batch cell line identification based on brightfield images. In contrast to previous approaches that use domain adaptation to forcibly align source and target domains, or incorporate multiple batches during model training to alleviate batch effect influence, we propose a different strategy, addressing batch effects by identifying three specific forms. Concentrating on these forms allows us to develop tailored solutions that effectively manage each type, consequently enhancing cell line identification based on brightfield images. Specifically, to account for the impact of cell density on cell images, we introduce a patch selection method called cell cluster-level selection (CCS), which identifies significant cell patches for downstream analysis. We also employ self-supervised learning (SSL) to extract robust patch representations that account for variations in image quality.  Finally, we present to use multiple instance learning (MIL) techniques to aggregate patch-level representations for cell line classification and propose a time-series segment sampling (TSS) module to address the bias introduced by differences in incubation times among cell batches. Our contributions can be summarized as follows:
\begin{enumerate}
    \item{We propose a novel patch selection method, Cell Cluster-level Selection (CCS), to account for the influence of cell density on cell images. CCS operates in an unsupervised manner, diversifying and filtering high-similarity proposals using complementary similarity measures, thereby producing representative cell patches for downstream analysis. We also introduce self-supervised learning (SSL) to extract robust embeddings from the patches, accounting for variations in image quality.}
    \item{To effectively utilize batch information and reduce prediction errors, we employ Multiple Instance Learning (MIL) for fusing patch-level representations in cell line classification tasks. In order to address the bias introduced by disparities in incubation times across cell batches, we propose a novel time-series segment sampling (TSS) module. This module enhances the feature-learning capabilities of the MIL aggregator by adaptively sampling patch embeddings, utilizing the available incubation timestamp information. Consequently, the MIL aggregator is better equipped to capture the dynamic changes and variations in cellular characteristics over time. }
    \item{We construct a large-scale dataset of 165190 brightfield cell images from 32 cell lines across 93 experimental batches, collected separately in practical biological experiments. This diversity allows for evaluating our method's effectiveness in addressing real-world batch effects in cell line identification. Through extensive experiments, we compare our approach to related works (e.g., domain adaptation, MIL) and validate each module using ablation studies. The results showcase the efficacy of our proposed method in identifying cell lines while addressing batch effects. To the best of our knowledge, this is the first work to tackle the problem of cross-batch cell line identification based on brightfield images.}
\end{enumerate}
The rest of this paper is structured as follows. We first present a comprehensive review of related works in image-based cell profiling for cell line identification, domain adaptation for mitigating batch effects, multiple instance learning for feature fusion, and point out how CLANet differs from them in Section \ref{sec:related_work}. In Section \ref{sec:dataset}, we introduce our experimental datasets. In Section \ref{sec:methodology}, we detail our proposed CLANet and present three distinct forms of batch effects with their tailored solutions. Experimental setups and results are presented in Section \ref{sec:experimetns} and \ref{sec:resutls}, respectively. Finally, we conclude this paper in Section \ref{sec:conclusion}.

\section{Related Work}
\label{sec:related_work}
\subsection{Image-based cell profiling for cell line identification}
Image-based cell profiling is a powerful technique that employs high-throughput imaging and computational analysis to study cellular phenotypes and morphological features for cell line identification. Various studies have explored different approaches to achieve accurate cell line identification using image-based cell profiling. \cite{belashov2021machine} constructed manual features (e.g. cell eccentricity, cell solidity) and employed three traditional ML classifiers (SVM, KNN, Adaboost) to classify cell phase images into three cell types (A549, HeLa, 3T3 cell lines) with three states (apoptosis, necrosis, live). \cite{mzurikwao2020towards} proposed identifying four cancer cell lines (COLO 704, EFO-21, EFO-27, UKF-NB-3) and their sublines adapted to anti-cancer drugs using Inception ResNet in an end-to-end manner. \cite{wang2020artificial} customized a bilinear CNN to identify seven pure cell lines and presented a segmentation network, DilatedNet, to analyze cross-contamination at the single-cell level. Most recently, \cite{tong2022automated} proposed a multi-task framework to identify thirty cell lines from brightfield cell images and predict the duration of how long cell lines have been incubated simultaneously. 

However, the generalizability of the previous findings for cross-batch cell line identification remains unclear, as their training and test data originate from the same batches of cells. To be suitable for standard laboratory practice, models should be able to differentiate between cell lines from distinct batch cultures.
\subsection{Domain adaptation for mitigating batch effects}
Batch effects, akin to the domain shift in machine learning, involve changes in data distribution between source and target domains, complicating model application across domains. Domain adaptation (DA) techniques address batch effects by transferring knowledge between domains. \cite{ando2017improving} employed a deep metric network for image representations and introduced the Typical Variation Normalization (TVN) method to enhance cell phenotype representations by removing nuisance variation effects. This approach improved drug classification by similar molecular mechanisms based on fluorescent cell images. \cite{cross2022self} proposed a weakly supervised self-supervised learning framework that effectively leverages compound information to refine phenotypical representations between source and target domains, achieving new state-of-the-art results on the BBBC021 dataset. \cite{sypetkowski2023rxrx1} employed Adaptive Batch Normalization (AdaBN), which modifies standard batch normalization layers to use statistics from individual domain distributions rather than the entire dataset, achieving good performance in classifying genetic perturbations based on fluorescent cell images derived from cell painting assays. \cite{jin2020minimum} proposed a loss function, termed "minimum class confusion", to handle four existing scenarios in domain adaptation - Closed-Set, Partial-Set, Multi-Source, and Multi-Target - with modifications.

Existing methods rely on using multiple data batches per class during model training or enforcing alignment between the source and target domains. However, these approaches may prove impractical due to unresolved issues concerning the required number of batches and associated collection processes. Moreover, cell culture data for cell line authentication pose a distinct challenge, as individual batches are collected separately, leading to a multi-source and multi-target problem. Additionally, cell line identification falls within the scope of fine-grained categorization, causing large intra-class variations and minimal inter-class variations between cell images. Enforcing the alignment of source and target domains under these conditions could introduce potential risks. To overcome these limitations, we present novel insights by categorizing batch effects into three distinct forms: cell density, image quality, and incubation times. We subsequently propose tailored solutions for each form, ultimately enhancing cell line identification even when relying on a single data batch per class during model training.

\subsection{Multiple instances learning for feature fusion}
Multiple Instance Learning (MIL) is a weakly supervised learning approach in which training instances are organized into sets, known as bags, with a label assigned to the entire bag. MIL can fuse the features of multiple instances within a bag, enabling the model to learn from aggregated information and capture the most relevant and discriminative patterns from the dataset. \cite{ilse2018attention} proposed an attention-based pooling for MIL in an end-to-end fashion, achieving impressive results on the cancer region detection in histopathology. \cite{li2021dual} proposed a dubbed dual-stream multiple instance learning network (DSMIL) to jointly learn multi-scale whole slide image features. \cite{chikontwe2021dual} integrated contrastive learning with the attention-based MIL aggregator to fuse CT slides features for Covid-19 classification. \cite{hashimoto2020multi} combined domain adaptation and MIL for the classification of malignant lymphoma in histopathology slides.

In this study, we apply the MIL aggregator to aggregate patch-level features from cell image sequences for cell line classification. To tackle the bias introduced by variations in incubation times among cell batches, we propose a Time-Series Segment Sampling (TSS) module to enhance the MIL aggregator's feature learning capabilities.


\section{Dataset}
\label{sec:dataset}
\begin{figure}
    \centering
    \includegraphics[width=0.5\textwidth]{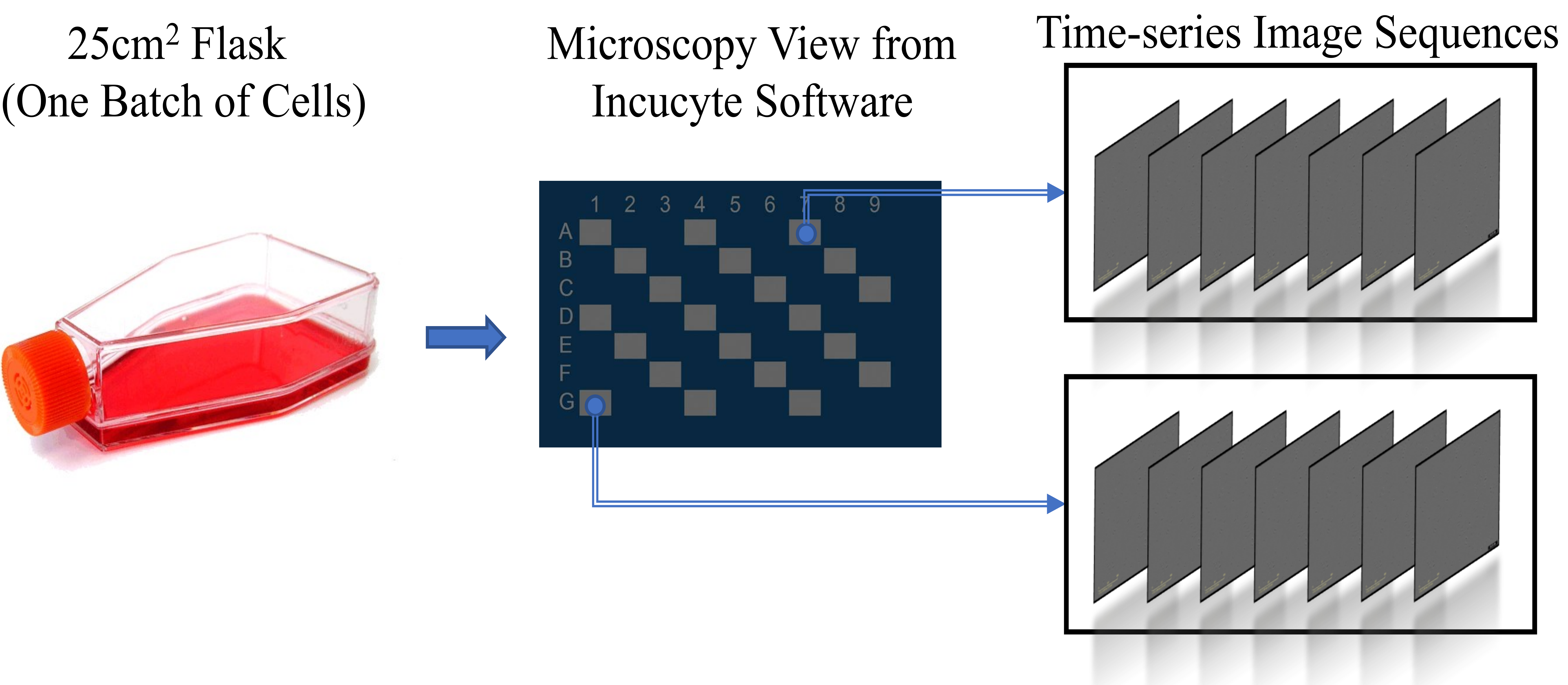}
    \caption{The workflow of the data collection. Cells are seeded in a 25$cm^{2}$ flask. The microscopy view of the flask is captured by the Incucyte software. Time-series image sequences can be exported from different locations (grey areas) across the flask.}
    \label{fig:data_collection}
\end{figure}

In this study, we have curated a rich dataset comprising sample images from the registered cell lines in the AstraZeneca Global Cell Bank (AZGCB). Our focus is predominantly on commonly used cancer cell lines, determined based on the frequency of requests for these cell lines. The dataset consists of 165190 brightfield images of 32 different cell lines across 93 experimental batches, as listed in Supplementary Table 1. All base medium was supplemented with 10$\%$ Foetal Bovine Serum (Sigma) and 1$\times$ GlutaMAX (Gibco) unless otherwise stated. Cells were thawed and seeded into a 25$cm^{2}$ flask (Corning) at a density of 0.5-2$\times 10^{6}$ cells per flask. The flasks were then placed in the Incucyte S3 system(Essen Bioscience, Sartorius), and brightfield images were collected from different locations across the flask every 1-8h over a time period between 3 and 18 days. An example of the data collection is depicted in Fig. \ref{fig:data_collection}. All images were exported for analysis as JPEG format with 1408$\times$1040 size (96$\times$96dpi).

Each data batch in this dataset corresponds to a specific cell line that was incubated during a unique biological experiment. The collected cell images formed part of the standard quality control process carried out within the AZ cell culture and banking team. Given the inherent variability in growth rates across cell lines, as well as differences in seeding density, IncuCyte usage, and incubation times, there are noticeable variations across the data batches. The diversity of this dataset allows us to assess the efficacy of our proposed method in addressing real-world batch effects in cell line identification. This is achieved by utilizing some data batches for model training, while reserving other batches for performance testing (details in Section \ref{sec:experimetns}). A summary of the dataset distribution is provided in Table \ref{tab:dataset_statis}.

\begin{table}[!htp]
 \centering
 \caption{Statistics of the AstraZeneca Global Cell Bank - Brightfield Imaging Dataset (AZGCB-BFID).}
 \begin{tabular}{cccccc}\toprule

             & Cell Lines & Batches & Image Sequences & Images    \\\midrule
        &32         & 93      & 2053            & 165190   \\
    
    \bottomrule
 \end{tabular}

 \label{tab:dataset_statis}
\end{table}

\section{Methodology}
\label{sec:methodology}
\begin{figure*}
    \centering
    \includegraphics[width = 1\textwidth]{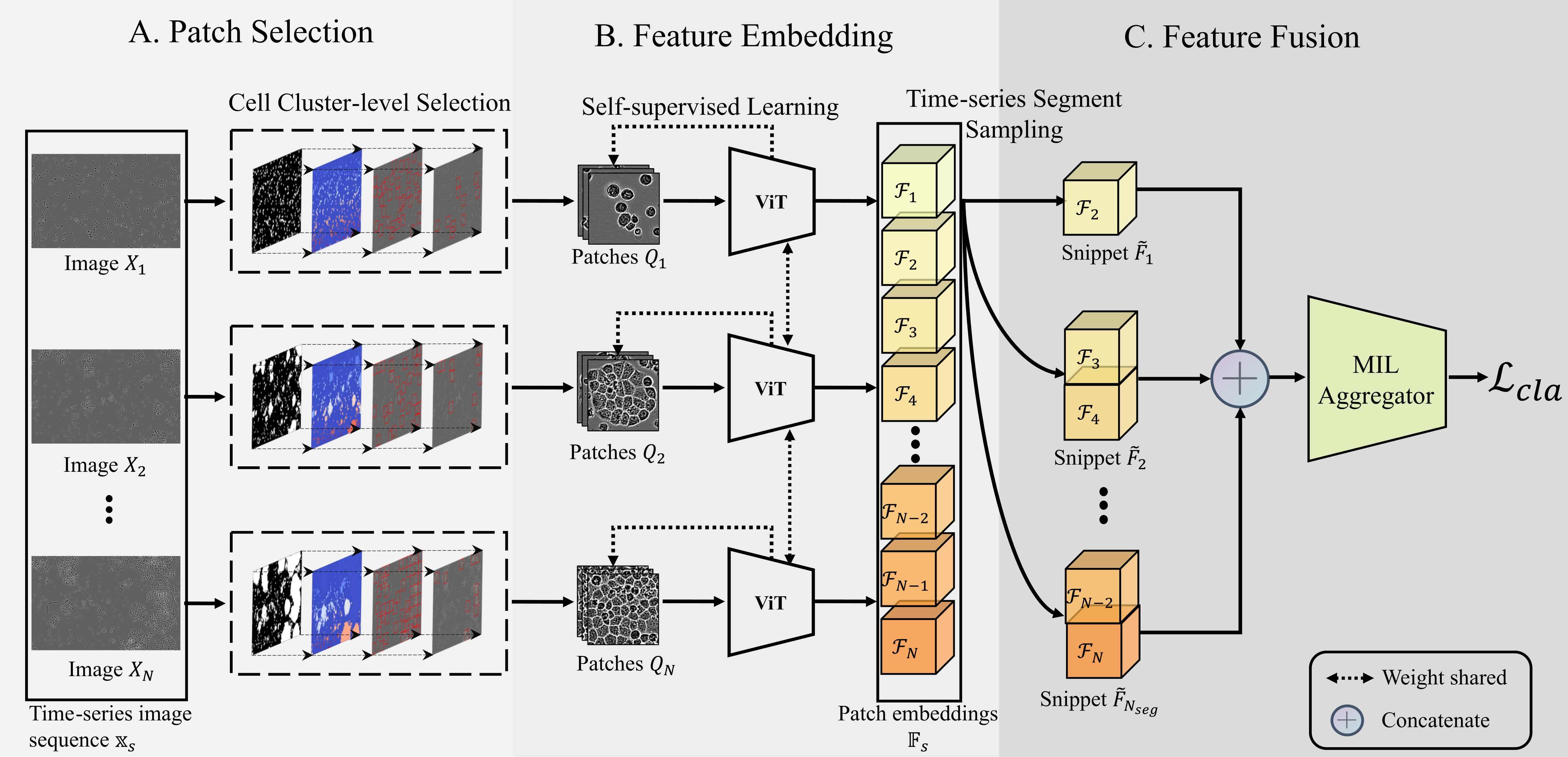}
    \caption{Pipeline of our proposed CLANet. A time-series cell image sequence $\mathbb{X}_{s}$ is obtained from a single microscopy location within a flask. Each cell image $X_{n}$ undergoes the Cell Cluster-level Selection to generate patches $Q_{n}$. Patch embeddings are extracted from patches using self-supervised learning, forming the patch embedding sequence $\mathbb{F}_{s}$. During training, the Time-series Segment Sampling is applied to sample the patch embedding sequence into several snippets, which are then fed into a multiple instance learning (MIL) aggregator to compute the classification loss $\mathcal{L}_{cla}$. During the inference stage, the complete embedding sequence is directly passed into the MIL aggregator to obtain the predicted label.}
    \label{fig:framework}
\end{figure*}

Our proposed framework comprises three stages (Fig. \ref{fig:framework}):  (1) extracting significant cell patches from the time-series image sequence; (2) self-supervised learning to learn and extract feature embeddings from the extracted patches; (3) feature fusion using the MIL aggregator for predicting cell line identity. 

In this section, we identify three special forms of batch effects and introduce our proposed tailored solutions for each challenge. To formulate our algorithm, we define our dataset as $\left\{(\mathbb{X}_{s}, \mathbb{Y}_{s})\right\}_{s=1}^{S}$, where $\mathbb{X}_{s}$ represents an image sequence obtained from a single microscopy location within a flask, and $\mathbb{Y}_{s}\in(1, C)$ denotes the class label, with C being the total number of classes. The image sequence $\mathbb{X}_{s}=\left\{X_{n}\right\}_{n=1}^{N}$ comprises $N$ cell images, each have a fixed size $W\times H$. An incubation timestamp set is defined as $\mathbb{T}_{s}=\left\{T_{n}\right\}_{n=1}^{N}$, corresponding to each cell image.

\subsection{Cell Cluster-level Selection for patch extraction}
\label{ccp}
\subsubsection{Problem Formulation}
\begin{figure}[h]
    \includegraphics[width=0.5\textwidth]{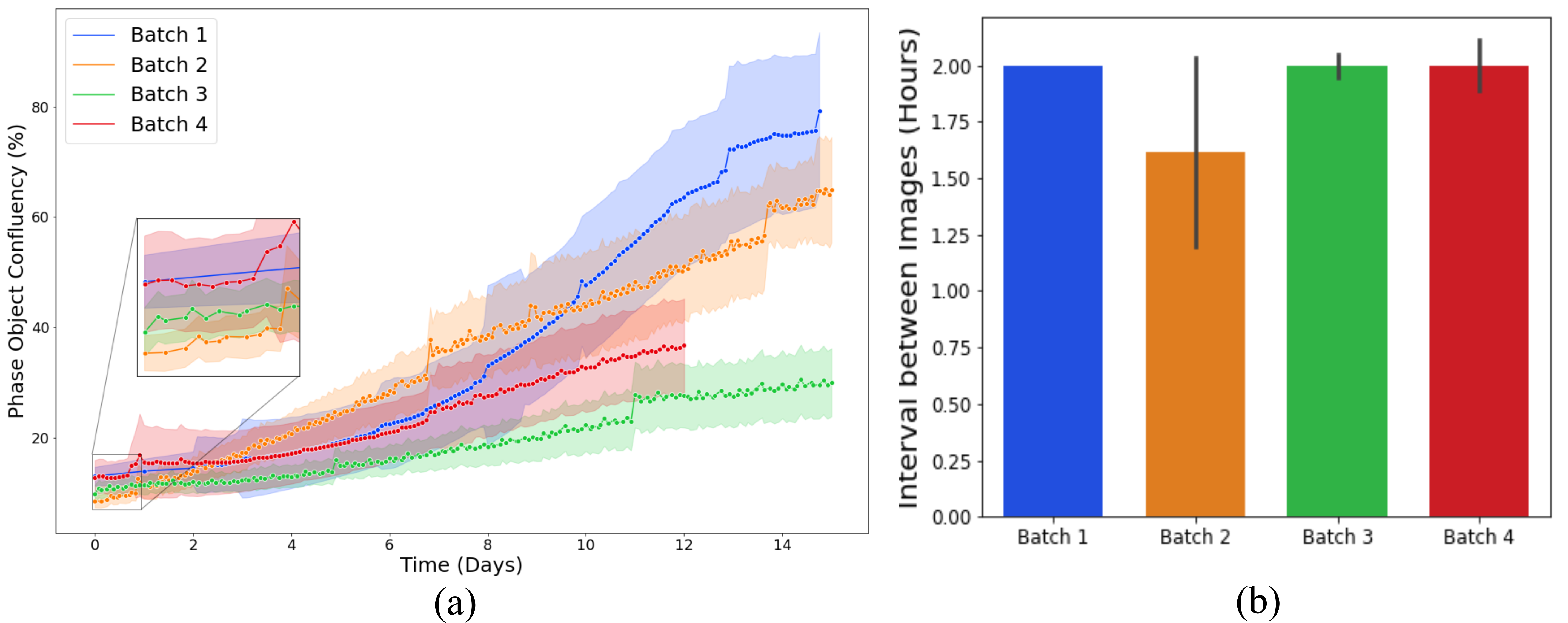}
    \caption{Batch-to-batch Variations in four batches of data from the CAMA1 cell line. (a) Variations in cell confluency over time. This figure illustrates the changes in cell confluency over time for four distinct batches. Each curve, accompanied by a shaded band, represents the mean and standard deviation of cell confluency values for image sequences within the same batch. (b) Variations in the average time interval ($\pm$ standard deviation) between cell images in the four batches.}
    \label{fig:batch_vari_forms}
\end{figure}

Cell confluency is commonly expressed as a percentage, indicating the fraction of the culture dish or flask covered by adherent cells. As depicted in Fig. \ref{fig:batch_vari_forms}(a), the 4 batches of data have different initial confluencies, and their growth curves exhibit different rates of change. The variation in confluencies leads to the collection of cell images with different cell densities, as seen in the images of batch 1 and batch 3 in Fig. \ref{fig:batch_effect}. Therefore, we propose that cell confluency is one of the sources of batch effects in the dataset.

Processing cell images at the single-cell level can be a way to reduce the impact of cell density on cell images. However, this approach has limitations due to the lack of single-cell contour annotations and the inter-class similarity of single cells between different cell types, which can limit its applicability in large-scale cell datasets (\cite{janssens2013charisma,yao2019cell}). Considering the biological characteristic of cells that they will attach and stick to the surface of the culture vessel in order to grow and proliferate, we propose to extract cell cluster-level patches as the representations of the original images. Different from previous attempts that split images into tiles and remove patches with low-density objects (\cite{li2021dual,su2022attention2majority}), we propose a novel patch selection method, namely cell cluster-level selection (CCS), to adaptively select discriminative patches from cell images. Since cell clusters have irregular contours, CCS first segments the image into connected regions, then diversifies and selects patches from the regions with high-density objects. This approach ensures that the selected patches capture representative features of the cell clusters and reduces the impact of cell density variations on downstream analysis.

\subsubsection{Proposed Method}
The initial step in CCS is generating binary masks for cell images to locate cell regions. Segmenting brightfield images presents challenges due to the complex interior structure of cells. Specifically, the pixel intensities of the cell membrane are brighter than the background, while the cytoplasm can be darker than the background. The presence of organelles further complicates the segmentation process. Popular pre-trained models, such as CellPose and SAM (\cite{pachitariu2022cellpose,kirillov2023segment}), cannot fully capture cell regions due to these challenges. Consequently, we propose a custom workflow for brightfield image segmentation (Supplementary Section 1.).

\RestyleAlgo{ruled}
\SetKwComment{Comment}{/* }{ */}
\begin{figure*}[h]
    \includegraphics[width=1\textwidth]{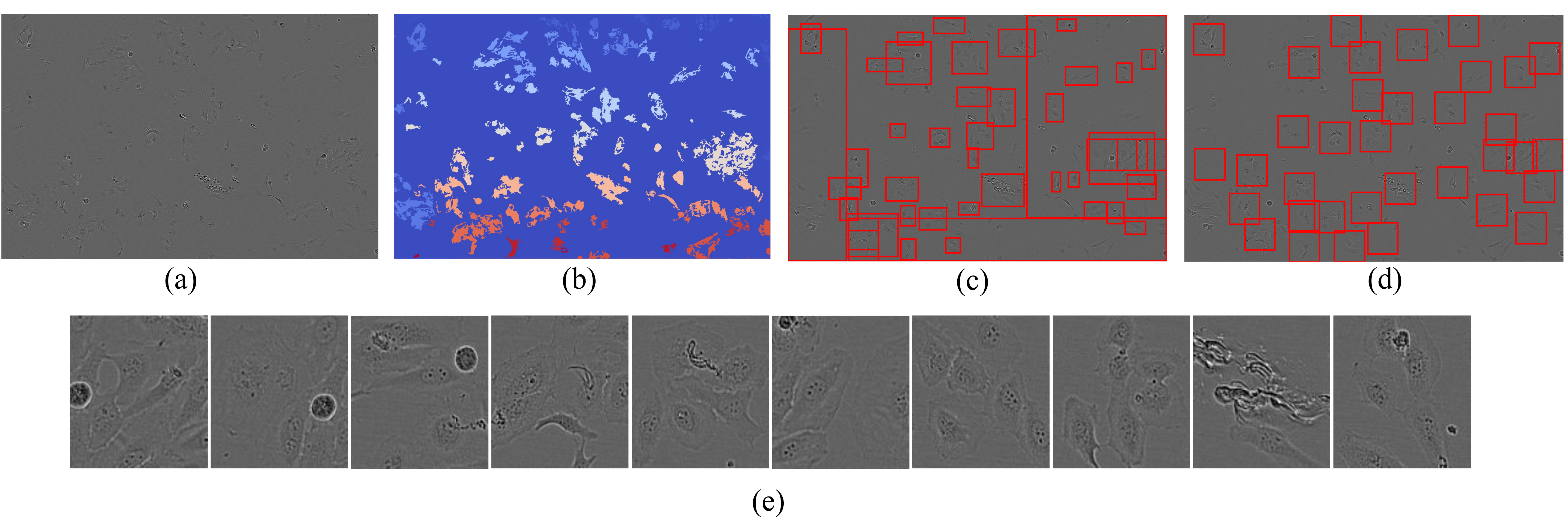}
    \caption{Outlines of the proposed Cell Cluster-level Selection (CCS). (a) a cell image. (b) connected region map. (c) initial bboxes. (d) candidate patches. (e) top 10 patches with the highest cell densities.}
    \label{fig:CCS}
\end{figure*}

Given a cell image $X_{n}$ with its binary mask $M_{n}$, we employ the Union-Find strategy (\cite{wu2005optimizing}) to label connected regions, resulting in disjoint subsets $\left\{m_{o}\right\}_{o=1}^{O} \gets M_{n}$ that correspond to separate cell regions in the mask, as depicted in Fig. \ref{fig:CCS}(b). We then add the initial bounding boxes (bboxes) $R_{n}=\left\{r_{o}\right\}_{o=1}^{O}$ along the contours of the connected regions (Fig. \ref{fig:CCS}(c)). While almost cell regions can be captured by bboxes of different sizes, our goal is to reduce the influence of cell density on cell images. To achieve this, we present capturing patches of fixed size while keeping the shape of the captured cell regions as uniform as possible. We correct the bboxes as follows:
\begin{equation}
\label{eqn:mass_center} 
\hat{R}_{n} \gets \left\{\begin{matrix*}[l]
\hat{R}_{n} \cup \text{MassCenter} (m_{o},W_{q},H_{q}),
\\ 
\hat{R}_{n} \cup \text{Tiles}(r_{o},W_{q},H_{q}), \ \ \ \  \text{if $\delta(r_{o})\geq 2W_{q}H_{q}$}
\end{matrix*}\right.
\end{equation}
where $W_{q}$ and $H_{q}$ are the predefined patch width and height. MassCenter$(\cdot)$ adjust the $r_{o}$ along the mass center of the cell cluster $m_{o}$ with the fixed size. When the area of the cell cluster $\delta(r_{o})$ is greater than or equal to double the predefined patch area, we further divide the region $r_{o}$ into tiles to fully capture the cells.

After creating the corrected bboxes, some overlap between them may occur. However, as each bbox may contain different cell morphology or texture information, it is not desirable to simply remove the overlapping components. Therefore, we draw inspiration from the selective search method (\cite{uijlings2013selective}) and measure the complementary similarity between the overlapping bboxes. We then remove the high-similarity components to increase the diversity of the bboxes. First, we define an overlapping pair as $(r_{a},r_{b})$ where the two bboxes have an overlapping region. The complementary similarity score $u(r_{a},r_{b})$ between the two bboxes is calculated as follows:
\begin{equation}
\label{eqn:similarity_score} 
u(r_{a},r_{b}) = \sum_{i=1}^{N_{ch}} \text{min}(ch_{a}^{(i)},ch_{b}^{(i)})+\sum_{i=1}^{N_{th}} \text{min}(th_{a}^{(i)},th_{b}^{(i)})+\frac{\left |r_{a}\bigcap r_{b}  \right |}{\left |r_{a}\bigcup r_{b}  \right |}
\end{equation}
The first term measures the color histogram similarity of the overlapping pair, where $ch_{a}$ and $ch_{b}$ represent the $N_{ch}$ color histograms of the two bboxes. The second term measures the texture histogram interaction between $th_{a}$ and $th_{b}$ using the Local Binary Pattern \cite{ojala2002multiresolution}. We further add the third term to calculate the interaction over union (IoU) of the two bboxes. By combining these three terms, we can effectively measure the similarity between the overlapping bboxes. To remove high-similarity components and increase the diversity of the bboxes, we use the following procedure:
\begin{equation}
\label{eqn:remove_sim} 
\hat{R}_{n} \gets \begin{cases}
			\hat{R}_{n}\setminus r_{a}, & \text{if $\delta(r_{a})<\delta(r_{b})$}\\
            \hat{R}_{n}\setminus r_{b}, & \text{otherwise}
		 \end{cases} \ \text{if}\  u(r_{a},r_{b})\geq \text{mean}(U_{n})
\end{equation}
We remove the low-density bbox if the similarity score between the bbox and its overlapping pair was greater than or equal to the average score across the score set $U_{n}$. We finally select and crop patches for downstream analysis from the top $K$ bboxes with the highest cell densities:
\begin{equation}
\label{eqn:top_k_patches} 
Q_{n}=\left\{q_{n,k} | \arg \max_{q_{n,k}\in \hat{R}_{n}\setminus \left \{q_{n,i}  \right \}_{i=0}^{k-1}}\ \delta(q_{n,k}),\ q_{n,0}\in\varnothing  \right\}_{k=1}^{K}
\end{equation}
as depicted in Fig. \ref{fig:CCS}(d)-(e). The algorithm of the proposed patch selection method is described in Algorithm \ref{algori:css}. 
\begin{algorithm}[!htb]
\caption{Cell Cluster-level Selection}
\label{algori:css}
 \textbf{Input:} Image $X_{n}$, Patch Number $K$, Patch Size $(W_{q},H_{q})$\\
  \textbf{Output:} $K$ Patches $Q_{n}$

       Create binary mask $M_{n}$ via Supplementary Algorithm 1\;
       Label connected regions $\left\{(m_{o})\right\}_{o=1}^{O} \gets M_{n}$\;
       Create initial bboxes $R_{n}=\left\{(r_{o})\right\}_{o=1}^{O}$\ using $\left\{(m_{o})\right\}_{o=1}^{O}$\;
       $\hat{R}_{n} = \varnothing $, $U_{n}= \varnothing$\;
       \For(\tcp*[f]{bbox correction}){$o \in O$} {
        $\hat{R}_{n} \gets \hat{R}_{n} \cup \text{MassCenter} (m_{o},W_{q},H_{q})$\;
        \If(\tcp*[f]{cell area $\delta(r_{o})$}){$\delta(r_{o})\geq 2W_{q}H_{q}$}{
           $\hat{R}_{n} \gets \hat{R}_{n} \cup \text{Tiles}(r_{o},W_{q},H_{q})$ \; 
       }
       }
       
       \ForEach{overlapping pair $(r_{a}\in \hat{R}_{n},r_{b} \in \hat{R}_{n})$}{
        Compute similarity score $u(r_{a},r_{b})$ via Eqn.\ref{eqn:similarity_score}\;
        $U_{n} = U_{n}\cup u(r_{a},r_{b})$\;
       } 
       \For(\tcp*[f]{remove high similarity components}){$u(r_{a},r_{b}) \in U_{n}$}{
       \If{$u(r_{a},r_{b})\geq \text{mean}(U_{n})$}{
       \eIf{$\delta(r_{a})<\delta(r_{b})$}{
        $\hat{R}=\hat{R}\setminus r_{a}$\;
       }{
       $\hat{R}=\hat{R}\setminus r_{b}$\;
       }
            
       }
       }
        
        \Return $Q_{n}=\left\{q_{n,k} | \arg \max_{q_{n,k}\in \hat{R}_{n}\setminus \left \{q_{n,i}  \right \}_{i=0}^{k-1}}\ \delta(q_{n,k}),\ q_{n,0}\in\varnothing  \right\}_{k=1}^{K}$
\end{algorithm}
\subsection{Self-supervised Learning for feature embeddings}
\label{ssl}
\subsubsection{Problem Formulation}
Image quality in cellular studies is heavily influenced by the microscopy system employed for acquisition. As shown in Fig. \ref{fig:batch_effect}, variations in contrast, brightness, and saturation may occur when different systems are used, leading to discrepancies in image quality. Consequently, this variation is recognized as a source of batch variation. Given that, in the previous step, we extracted a diverse array of patches to represent the original cell image, intending to minimize the influence of cell density. It is crucial to extract patch features and address image quality variations at this stage to ensure robust downstream analysis and reliable results.

To tackle these challenges, we advocate incorporating self-supervised learning (SSL) into our analysis pipeline. In our context, the goal is to leverage SSL to extract meaningful features from the inherent structure of cell patches, circumventing reliance on annotated cell class labels. By adopting this approach, we aim to mitigate the risk of biased predictions arising from batch effects, thereby ensuring more robust and generalizable outcomes in the final prediction.
\subsubsection{Proposed Method}
We consider Dino \cite{caron2021emerging}, a state-of-the-art (SOTA) SSL framework that enables robust representations to be learned for unlabeled data. Dino makes use of a self-distillation loss function that encourages the student model to produce similar representations for augmented versions of the same input image while making representations for different images more distinct.

In this section, we implement Dino to acquire the feature embeddings from the cell patches. For each patch $q_{n,k}$, a set of views is generated using a multi-crop strategy (\cite{caron2020unsupervised}), which contains several global ${q_{n,k}}'$ and local crops ${q_{n,k}}''$. To enable the model to mitigate the effects of batch variation in image quality, we further introduce random augmentations, such as random contrast, brightness, and vertical flip, into the crops. 
We define a teacher network as $G_{tm}(\cdot;\theta_{tm})$ with its trainable parameters $\theta_{tm}$, and a student network is denoted as $G_{sm}(\cdot;\theta_{sm})$. The probabilities $P_{tm}(\cdot)$ and $P_{sm}(\cdot)$ are obtained by normalizing the outputs of the two networks with a softmax function. The training task of the student network is formulated as the following minimization problem:
\begin{equation}
\label{eqn:ssl} 
\begin{aligned}
\hat{\theta}_{sm}\gets  &\arg \min_{\theta_{sm}} \sum_{i_g=1}^{N_{global}} \sum_{i_l=1}^{N_{local}}  P_{tm}({q^{(i_{g})}_{n,k}}';\theta_{tm})\log P_{sm}({q^{(i_{l})}_{n,k}}'';\theta_{sm}) \\ &+ \sum_{i_g=1}^{N_{global}}\sum_{\substack{j_g=1\\j_g\neq i_{g}}}^{N_{global}}  P_{tm}({q^{(i_{g})}_{n,k}}';\theta_{tm})\log P_{sm}({q^{(j_{g})}_{n,k}}';\theta_{sm})
\end{aligned}
\end{equation}
where $N_{global}$ and $N_{local}$ represent the number of global and local views separately. Teacher weights $\theta_{tm}$ are frozen during each epoch of student training and updated iteratively with an exponential moving average of the student weights: $\hat{\theta}_{tm} \gets \lambda \theta_{tm} + (1-\lambda)\theta_{sm}$, where $\lambda$ is the momentum parameter. After the training converges, a patch embedding list $\mathcal{F}_{n}=\left \{f_{n,k}  \right \}_{k=1}^{K}$ can be obtained by passing the cell patches into the teacher network $f_{n,k}=G_{tm}(q_{n,k};\theta_{tm})$. $\mathcal{F}_{n} \in \mathbb{R}^{K \times D} $ where $D$ is the dimension of the output embeddings.

\subsection{Time-series Segment Sampling for Multiple Instance Learning}
\label{mil}
\subsubsection{Problem Formulation}
The third batch variation arises from differences in incubation times, referring to the duration for which cells are cultured in a flask. As illustrated in Fig. \ref{fig:batch_vari_forms}(a)-(b), varying incubation times may be necessary for different batches of cells, even when using the same cell line, due to factors such as temperature, humidity, or variability in initial cell seeding density. Furthermore, the interval between time-lapse image acquisitions can introduce bias as a result of discrepancies in the practices of individual biologists. These variations lead to the collection of different numbers of cell images with varying time intervals, potentially affecting the consistency and reliability of the subsequent analyses. 

 To address the challenge of varying incubation times and inconsistent sample sizes, we propose the use of multiple instance learning (MIL) to fuse the feature embeddings of patches for the final prediction of cell identity. Distinct from the standard MIL setting, where a bag is labeled positive if it contains at least one positive instance and negative if all its instances are negative (\cite{li2021dual,wang2023deep}), we formulate the MIL problem as a multi-class classification problem. Considering that different batches may include varying numbers of the time-series image sequences, we treat the image sequence as the unit for MIL (i.e. bag). Thus, we can acquire a time-series patch embedding sequence $\mathbb{F}_{s} = \left \{ \mathcal{F}_{n} \right \}_{n=1}^{N}$ from the above steps. The multi-class MIL problem is defined as follows:
\begin{equation}
\label{eqn:mil_formulation} 
P_{mm}(\mathbb{F}_{s}) = \text{Softmax}(G_{mm}(\mathcal{F}_{1},...,\mathcal{F}_{N})), \ \ P_{mm}(\mathbb{F}_{s})\in \mathbb{R}^{C}
\end{equation}
where $G_{mm}(\cdot)$ is an aggregation operator and $P_{mm}(\mathbb{F}_{s})$ is the bag-level probability.

\begin{figure}
    \includegraphics[width=0.49\textwidth]{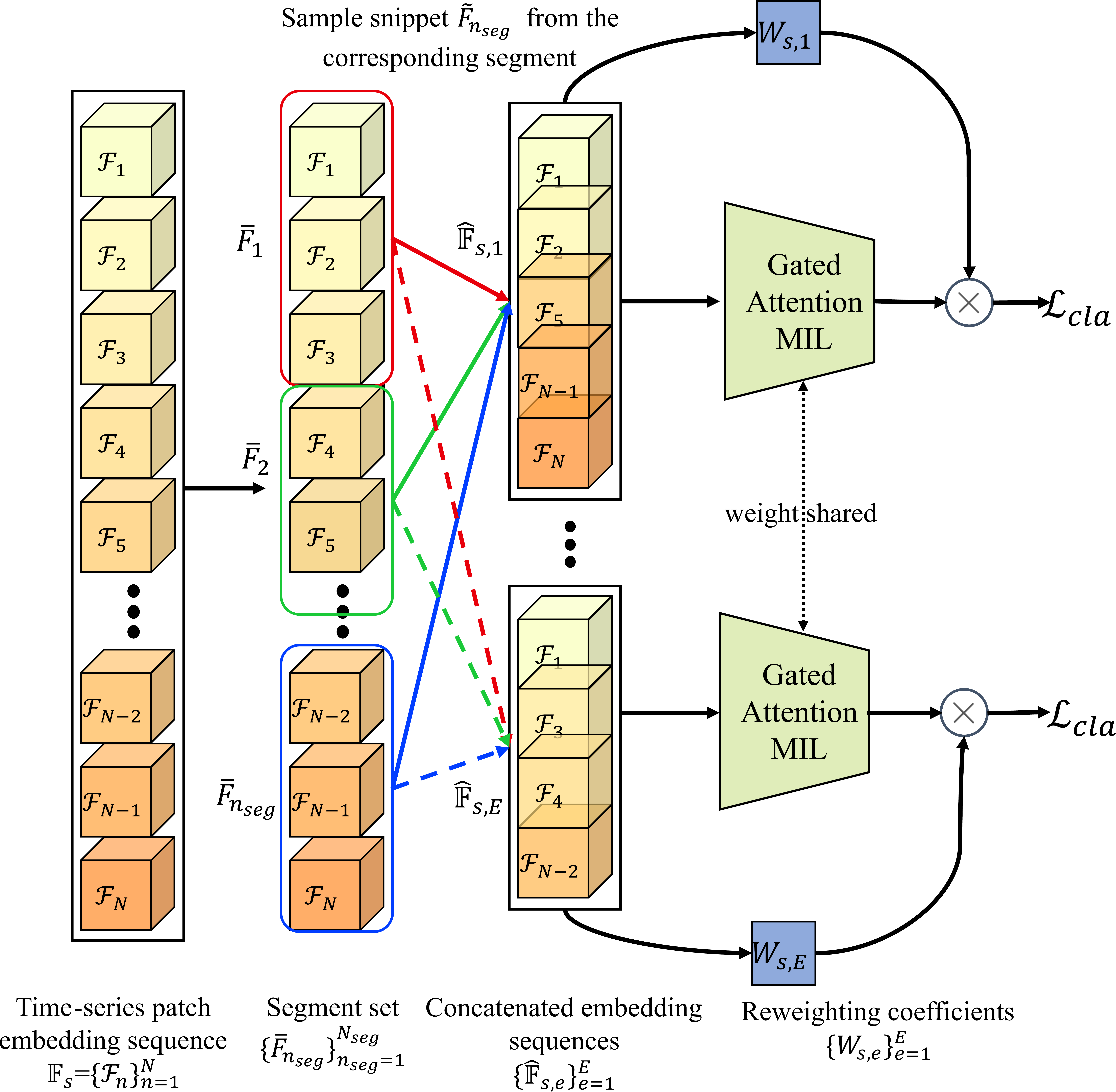}
    \caption{Illustration of the proposed Time-series Segment Sampling (TSS) for Multiple Instance Learning (MIL). A complete time-series patch embedding sequence $\mathbb{F}_{s}$ will be partitioned into non-overlapping segments $\left \{ \bar{F}_{n_{seg}} \right \}^{N_{seg}}_{n_{seg}=1}$. A snippet, $\tilde{F}_{n_{seg}}$, is randomly sampled from each corresponding segment. In each training epoch, a constructed embedding sequence $\hat{\mathbb{F}}_{s,e}$ will to fed into the gated Attention MIL and compute a classification loss $\mathcal{L}_{cla}$ with the weight coefficient $W_{s,e}$.}
    \label{fig:TSS_MIL}
\end{figure}

Although the MIL aggregator effectively aggregates information and discerns relevant patterns for specific cell lines from the embedding sequence, its performance in identifying cell lines may be hindered due to the variability in cell image numbers and time intervals. Drawing inspiration from the Temporal Segment Network (TSN) (\cite{wang2018temporal}, we propose a novel module, Time-Series Segment Sampling (TSS), to bolster the feature learning capabilities of the MIL aggregator. In the TSS module, a Gaussian distribution is constructed based on the incubation timestamp information of cell images, to adaptively determine segment lengths and partition image sequences accordingly.  Subsequently, snippets are sampled from each segment, culminating in a new embedding sequence. A gated attention MIL aggregator is employed to fuse the sequence of embeddings, effectively learning the temporal dynamics and variations in cellular characteristics. An illustration of the proposed TSS module is shown in Fig. \ref{fig:TSS_MIL}. 

\subsubsection{Proposed Method}
The initial step of TSS involves constructing a Gaussian distribution to adaptively sample various time intervals. This approach aims to account for the interval bias between cell images originating from different batches. In cases where an embedding sequence with a shorter time interval (e.g., 2h) is present, sparse sampling can be applied to approximate a sequence with longer time intervals (e.g., 4h, 8h). Conversely, retaining the original sequence is preferable to avoid the potential introduction of additional noise. The choice to employ a Gaussian distribution for sampling time intervals is based on the observation that longer time intervals are expected to be relatively rare in practice. As a result, the distribution's characteristics enable more robust handling of varying time intervals while minimizing the impact of extreme values. Given a time-series patch embedding sequence $\mathbb{F}_{s} = \left \{ \mathcal{F}_{n} \right \}_{n=1}^{N}$ with its corresponding timestamp set $\mathbb{T}_{s}= \left\{T_{n}\right\}_{n=1}^{N}$, a Gaussian distribution $\mathcal{N}(\cdot)$ can be constructed as follows:
\begin{equation}
\label{eqn:gaussian_mu} 
\mathcal{N}(\mu(\mathbb{T}_{s}),\sigma^{2}) \gets\left\{\begin{matrix*}[l]
\mu(\mathbb{T}_{s}) = \left |  \frac{\sum_{i=1}^{N-1}(T_{i+1}-T_{i})}{N-1}\right | \\ 
\sigma = \text{std}(\left \{ \mu(\mathbb{T}_{s}) \right \}_{s=1}^{S})
\end{matrix*}\right. ,\ \mu(\mathbb{T}_{s}) \in \mathbb{Z}
\end{equation}
where std$(\cdot)$ represents the standard deviation of the expected time interval set. 

The sampling of time intervals is integrated with epoch training, with the goal of leveraging the Gaussian distribution to globally control the MIL model's training. Define the number of training epochs as $E$, a set of the time intervals is generated from the Gaussian distribution as follows:
\begin{equation}
\label{eqn:gaussian_interval_set} 
\left \{ \tilde{T}_{e}| \tilde{T}_{e} \sim \mathcal{N}(\mu(\mathbb{T}_{s}),\sigma^{2}),\ \tilde{T}_{e} \in \left \{ \mu(\mathbb{T}_{s}) \right \}\right \}_{e=1}^{E}
\end{equation}
For each training epoch, a sampled time interval $\tilde{T}_{e}$ is partition the embedding sequence into non-overlapping segments $\left \{ \bar{F}_{n_{seg}} \right \}^{N_{seg}}_{n_{seg}=1}$, where the segment length is $\left\lfloor\frac{\tilde{T}_{e}}{\mu(\mathbb{T}_{s})} \right\rfloor \in \mathbb{Z}$ and $N_{seg}=\left\lceil\frac{T_{N}}{\tilde{T}_{e}}+1\right\rceil \in \mathbb{Z}$. A snippet, $\tilde{F}_{n_{seg}}$, is randomly sampled from each corresponding segment. This operation aims to prevent abrupt sampling jumps and the omission of pertinent information, ensuring that the model can effectively capture the diverse temporal patterns present in the data. A new embedding sequence $\hat{\mathbb{F}}_{s,e}$ is formed as follows:
\begin{equation}
\label{eqn:concat_snippet} 
\hat{\mathbb{F}}_{s,e} = \begin{cases}
			\text{Concat}(\tilde{F}_{1}, ..., \tilde{F}_{N_{seg}} ), & \text{if $\tilde{T}_{e}\leq \mu(\mathbb{T}_{s})$}\\
            \left\{\mathcal{F}_{1},...,\mathcal{F}_{N}\right\}, & \text{otherwise}
		 \end{cases}
\end{equation}
where Concat$(\cdot)$ represents the concatenate operation. When $\tilde{T}_{e} \leq \mu(\mathbb{T}_{s})$, it is preferable to retain the original sequence to prevent the potential introduction of additional noise. This setting is also appropriate for sequences with long time intervals (e.g., 8h), as these sequences do not display nearly continuous temporal phenotypic features. In such cases, the objective is to enable the model to learn from the original sequence as extensively as possible.

A gated attention MIL aggregator (\cite{ilse2018attention}) is utilized to fuse the embeddings $\hat{\mathbb{F}}_{s,e}$, generating sequence-level representation for predicting cell line identity. Denote the length of $\hat{\mathbb{F}}_{s,e}$ as $V$, the sequence-level representation is expressed as:
\begin{equation}
\label{eqn:gated_attention} 
G_{mm}(\hat{\mathbb{F}}_{s,e} ) = \sum_{v=1}^{V}\sum_{k=1}^{K}A_{v,k}f_{v,k} 
\end{equation}
where:
\begin{equation}
\label{eqn:attention_scores} 
A_{v,k} = \frac{\text{exp}\left\{\psi^\top(\text{tanh}(\vartheta f_{v,k}^\top )\odot \text{sigm}(\phi f_{v,k}^\top  ))\right\}}{\sum_{i=1}^{V}\sum_{j=1}^{K}\text{exp}\left\{\psi ^\top(\text{tanh}(\vartheta f_{i,j}^\top  )\odot \text{sigm}(\phi f_{i,j}^\top  ))\right\}} 
\end{equation}
Here, $\psi $, $\vartheta$ and $\phi$ represent the parameters of the MIL aggregator. $\odot$ denotes element-wise wise multiplication, while tanh$(\cdot)$ and sigm$(\cdot)$ represent the tangent and sigmoid non-linearities, repectively.

Considering the potential risks introduced by the sampling procedure, which may discard some discriminative embeddings, we further propose a reweighting mechanism to balance the trade-off between preserving complete information and adapting to varying time intervals. This weighting coefficient is expressed as follows:
\begin{equation}
\label{eqn:weight_seq} 
W_{s,e} = \alpha_{1}\cdot \frac{V}{N}   +\alpha_{2}\cdot \frac{\mu(\hat{\mathbb{T}}_{s,e})-\mu(\mathbb{T}_{s})}{\tilde{T}_{e}-\mu(\mathbb{T}_{s})}
\end{equation} 
where $\hat{\mathbb{T}}_{s,e}$ represents the corresponding time set of $\hat{\mathbb{F}}_{s,e}$, and $(\alpha_{1},\alpha_{2})\in\left\{0,1\right\}$ indicate whether the respective term is used or not. Our classification loss function is defined as follows:
\begin{equation}
\label{eqn:CE_1} 
\mathcal{L}_{cla} = W_{s,e} \cdot \mathcal{L}_{ce}(\mathbb{Y}_{s},P_{mm}(\hat{\mathbb{F}}_{s,e}))
\end{equation}
where $\mathcal{L}_{ce}$ denotes the cross entropy loss.

An additional challenge observed in previous MIL-related works \cite{ilse2018attention,li2021dual} is the use of mini-batch size 1 for training the MIL model due to varying bag sizes. This results in unstable convergence of the training loss. To address this issue, we introduce gradient accumulation, a technique that accumulates gradients over several batches before updating the model's parameters. By doing so, we effectively simulate training with larger batch sizes and the gradient of $\mathcal{L}_{cla}$ with respect to all of the training parameters $\theta_{mm}^{(all)}$ is expressed as follows:
\begin{equation}
\label{eqn:CE_final} 
\frac{\partial \mathcal{L}_{cla}}{\partial \theta_{mm}^{(all)}}= \sum_{b=1}^{B}(\frac{W_{b,e}}{\sum_{i=1}^{B}W_{i,e}}\cdot \frac{\partial \mathcal{L}_{ce}(\mathbb{Y}_{b},P_{mm}(\hat{\mathbb{F}}_{b,e}))}{\partial \theta_{mm}^{(all)}})
\end{equation}
where $B$ is the mini-batch size. During the inference stage, the complete patch embedding sequence will be passed into the MIL aggregator to obtain the predicted label. The entire process of the TSS module for MIL is described in Algorithm \ref{algori:tss}.

\begin{algorithm}[!htb]
\caption{Time-series Segment Sampling for Multiple Instance Learning}
\label{algori:tss}
 \textbf{Input:} Time-series patch embedding sequence $\mathbb{F}_{s} = \left \{ \mathcal{F}_{n} \right \}_{n=1}^{N}$, timestamp set $\mathbb{T}_{s}= \left\{T_{n}\right\}_{n=1}^{N}$, Reweighting terms $\alpha_{1}$ and $\alpha_{2}$, Epoch number $E$, Mini-batch size $B$, \\
       
       $\mu(\mathbb{T}_{s}) = \left |  \frac{\sum_{i=1}^{N-1}(T_{i+1}-T_{i})}{N-1}\right | $, $\sigma = \text{std}(\left \{ \mu(\mathbb{T}_{s}) \right \}_{s=1}^{S})$\;
       
       $\left \{ \tilde{T}_{e}| \tilde{T}_{e} \sim \mathcal{N}(\mu(\mathbb{T}_{s}),\sigma^{2}),\ \tilde{T}_{e} \in \left \{ \mu(\mathbb{T}_{s}) \right \}\right \}_{e=1}^{E}$\;
       \For(\tcp*[f]{epoch training}){$e\in E $}{
        
        Split the embedding sequence $\left \{ \mathcal{F}_{n} \right \}_{n=1}^{N}$ into segments $\left \{ \bar{F}_{n_{seg}} \right \}^{N_{seg}}_{n_{seg}=1}$ using the interval $\tilde{T}_{e}$\;
        Sample snippet $\left \{ \tilde{F}_{n_{seg}} \right \}^{N_{seg}}_{n_{seg}=1}$ from segments\;
        $\hat{\mathbb{F}}_{s,e} = \varnothing$\;
        \eIf{$\tilde{T}_{e}\leq \mu(\mathbb{T}_{s})$}{
        $\hat{\mathbb{F}}_{s,e} = \text{Concat}(\tilde{F}_{1}, ..., \tilde{F}_{N_{seg}} )$\;
       }{
       $\hat{\mathbb{F}}_{s,e} =\left\{\mathcal{F}_{1},...,\mathcal{F}_{N}\right\} $\;
       }
       Generate sequence-level representation $G_{mm}(\hat{\mathbb{F}}_{s,e} )$ via Eqns. \ref{eqn:gated_attention}-\ref{eqn:attention_scores}\;
       Compute the classification loss $\mathcal{L}_{cla}$ by Eqns. \ref{eqn:weight_seq}-\ref{eqn:CE_final}\;
       }
       
\end{algorithm}

\section{Experiments}
\label{sec:experimetns}
We demonstrate the effectiveness of the proposed CLANet for cross-batch cell line identification on the established AZGCB-BFID dataset. The details of the experimental settings are presented in the following subsections.
\subsection{Evaluation Tasks and Metrics}
To ensure a comprehensive evaluation, inspired by \cite{sypetkowski2023rxrx1}, we introduce two strategies for splitting the data into training and test sets.

\textbf{Batch-separated:} We randomly select 1 batch of data from each of the 32 cell classes as the training set, while the remaining 61 batches are assigned to the test set. In this way, the experimental batches in the test set differ from those in the training set, enabling the evaluation of cross-batch generalization (i.e. out-of-domain generalization) performance.

\textbf{Batch-stratified:} In this strategy, both the training and test sets contain data from all experimental batches. The data are split within each batch, forming the sets. The sizes of the training and test sets are made roughly the same as in the batch-separated split. This split strategy is used to explore the influence of batch effects on classification performance in comparison to other methods.

The random data split is performed 3 times for each of the above strategies separately. Since the input unit for the proposed method is the image sequence, the evaluation is conducted on two aspects: sequence-level (seq-level) and batch-level. The seq-level evaluation focuses on individual image sequences in relation to their respective cell line class labels, while the batch-level evaluation averages the results of all image sequences within the same batch. Top-1 accuracy and F1-score are utilized as criterias for assessing classification performance. The statistics of data splitting are presented in Table \ref{tab:dataset_split}.

\begin{table}[!htp]
 \centering
\footnotesize
 \caption{Statistics of the data splitting on the AZGCB-BFID dataset. The number of sequences may vary slightly during the random data split, as the sequence count in each batch is not equal.}
\begin{tabular}{*{5}{c}}
  \toprule
  \multirow{2}*{Splitting Strategy} & \multicolumn{2}{c}{Training} & \multicolumn{2}{c}{Test} \\
  \cmidrule(lr){2-3}\cmidrule(lr){4-5}
  & Batches &Sequences  & Batches &Sequences \\
  \midrule
  Batch-separated & 32 & 688$\sim $746 & 61 & 1307$\sim $1365 \\
  Batch-stratified & 93 &  729$\sim $763  & 93 & 1290$\sim $1324 \\
  \bottomrule
\end{tabular}

 \label{tab:dataset_split}
\end{table}

\begin{table*}

\scriptsize
 \caption{Comparison of our proposed framework with different models for classifying cell lines on the AZGCB-BFID dataset, reporting the results as [mean value $\pm$ standard deviation]. Each model was trained three times on both batch-separated and batch-stratified splits. }

\begin{tabular}{{lcccccccc}}
  \toprule
  
  Splitting Strategy & \multicolumn{4}{c}{Batch-separated} & \multicolumn{4}{c}{Batch-stratified}  \\
  \cmidrule(lr){2-5}\cmidrule(lr){6-9}\ 
  \multirow{2}*{Methods}&\multicolumn{2}{c}{Accuracy.\%}&\multicolumn{2}{c}{F1-score.\%}&\multicolumn{2}{c}{Accuracy.\%}&\multicolumn{2}{c}{F1-score.\%}\\
  \cmidrule(lr){2-3}\cmidrule(lr){4-5}\cmidrule(lr){6-7}\cmidrule(lr){8-9}\ 
  & Sequence-level &Batch-level  & Sequence-level &Batch-level& Sequence-level &Batch-level& Sequence-level &Batch-level \\
  \midrule
  ViT(w/ Images)  & 51.2$\pm$2.8 & 55.4$\pm$3.0 & 41.9$\pm$2.6 & 48.2$\pm$2.1& 98.7$\pm$0.5& 99.6$\pm$0.4& 98.1$\pm$0.7& 99.4$\pm$0.6 \\
  
  ViT(w/ CCS) & 70.2$\pm$4.1 & 72.0$\pm$3.3 & 57.1$\pm$3.1 & 60.6$\pm$2.4& 99.3$\pm$0.3& 99.7$\pm$0.3& 99.1$\pm$0.2& 99.5$\pm$0.5 \\
  \midrule

  UDMIL(\cite{wang2020ud})  & 73.0$\pm$3.4 & 77.4$\pm$4.6 & 65.1$\pm$4.2 & 68.3$\pm$4.8& 99.3$\pm$0.6& 99.7$\pm$0.2& 99.0$\pm$0.4& 99.4$\pm$0.4 \\
  
  DSMIL(\cite{li2021dual})  & 78.4$\pm$3.6 & 82.2$\pm$3.9 & 68.6$\pm$2.4 & 74.8$\pm$3.6& 99.1$\pm$0.7& 100.0$\pm$0.0& 99.3$\pm$0.4& 100.0$\pm$0.0 \\

  MCC(\cite{jin2020minimum})  & 68.9$\pm$4.2 & 69.9$\pm$2.6 & 55.1$\pm$3.5 & 59.1$\pm$1.8& 99.1$\pm$0.2& 99.6$\pm$0.2& 99.1$\pm$0.3& 99.5$\pm$0.3\\
  
  DAMIL(\cite{hashimoto2020multi})  & 74.7$\pm$4.2 & 76.8$\pm$3.6 & 66.4$\pm$2.4 & 67.8$\pm$2.3& 98.9$\pm$0.6& 99.4$\pm$0.4& 99.1$\pm$0.4& 99.3$\pm$0.4 \\
 
  CLANet(\textbf{ours})  & \textbf{89.1$\pm$2.2} & \textbf{90.7$\pm$2.0} & \textbf{83.7$\pm$4.2} & \textbf{87.0$\pm$4.8}& 99.9$\pm$0.1& 100.0$\pm$0.0& 99.9$\pm$0.1& 100.0$\pm$0.0 \\

  \bottomrule
\end{tabular}
\label{tab:comparison_methods}
\end{table*}

\begin{table*}
\centering

\scriptsize
 \caption{Ablation studies on the key components of the framework on the batch-separated task.}

\begin{tabular}{{lcccc}}
  \toprule

  \multirow{2}*{Methods}&\multicolumn{2}{c}{Accuracy.\%}&\multicolumn{2}{c}{F1-score.\%}\\
  \cmidrule(lr){2-3}\cmidrule(lr){4-5}\ 
  & Sequence-level &Batch-level  & Sequence-level &Batch-level \\
  \midrule
  ViT(w/ Images)+GAMIL  & 43.1$\pm$3.7 & 45.0$\pm$3.7 & 33.1$\pm$3.4 & 34.8$\pm$4.3  \\
  ViT(w/ Tiles)+GAMIL & 42.9$\pm$1.8 & 45.2$\pm$2.6 & 33.9$\pm$2.1 & 35.2$\pm$4.7  \\
  ViT(w/ CCS) & \textbf{70.2$\pm$4.1} & \textbf{72.0$\pm$3.3} & \textbf{57.1$\pm$3.1} & \textbf{60.6$\pm$2.4}  \\
   ViT(w/ CCS)+GAMIL & 57.4$\pm$6.7 & 62.4$\pm$5.9 & 47.1$\pm$4.5 & 53.0$\pm$4.7  \\
  \midrule
  ViT(w/ Images)+SSL+GAMIL  & 68.8$\pm$0.9 & 72.6$\pm$1.3 & 60.9$\pm$1.7 & 66.2$\pm$1.5  \\
  ViT(w/ Tiles)+SSL+GAMIL & 71.8$\pm$3.7 & 70.4$\pm$4.3 & 65.6$\pm$4.6 & 64.3$\pm$2.2  \\
  ViT(w/ CCS)+SSL &86.4$\pm$2.0 & 85.7$\pm$1.5 & 76.1$\pm$5.2 & 78.0$\pm$4.9  \\
  ViT(w/ CCS)+SSL+GAMIL & \textbf{87.0$\pm$2.7} & \textbf{86.8$\pm$3.6} & \textbf{80.9$\pm$3.6} & \textbf{84.2$\pm$2.9}  \\
   \midrule
   
  ViT(w/ CCS)+SSL+Max-Pooling & 77.7$\pm$4.6 & 79.6$\pm$7.7 & 68.0$\pm$5.5 & 72.7$\pm$6.1  \\
  ViT(w/ CCS)+SSL+Avg-Pooling & 84.8$\pm$2.1 & 84.9$\pm$3.3 & 78.0$\pm$5.0 & 79.8$\pm$4.1  \\
  ViT(w/ CCS)+SSL+TSS+GAMIL & \textbf{89.1$\pm$2.2} & \textbf{90.7$\pm$2.0} & \textbf{83.7$\pm$4.2} & \textbf{87.0$\pm$4.8}  \\

  \bottomrule
\end{tabular}
\label{tab:ablation_study}
\end{table*}

\subsection{Implementation Details}
Our cell line authentication framework consists of three components: Patch Selection (CCS), Feature Embedding (SSL), and Feature Fusion (TSS-MIL). Experiments are conducted using 2 NVIDIA Tesla V100 GPUs for hardware, and Python, Pytorch, and Scikit-Image for the software environment.

In the patch selection stage, we employ the CCS method to locate patch proposals within each cell image and subsequently crop the top $K$=10 patches with sizes of $(W_{q},H_{q})=$112$\times$112. During the feature embedding stage, we train Dino to obtain feature embeddings for all patches. We utilizes ViT-S/8 as the backbone of teacher and student models, and for each patch, 2 global 112$\times$112 random resize crops and 8 local 32$\times$32 random resize crops are performed, accompanied by random color jitter, vertical, and horizontal flip augmentations during training. The initial learning rate is set at $5\times10^{-5}$, and all models are initialized with weights pre-trained on the ImageNet dataset. The training process is executed over 200 epochs with a mini-batch size of 256. At the end of training, we extract an embedding of size D=1536 for each patch.

In the feature fusion stage, the maximum epoch $E$ for the training of MIL is set to 2000, ensuring that the Gaussian sampling can adequately fit the distribution. Both $\alpha_{1}$ and $\alpha_{2}$ are assigned a value of 1 to enable the functioning of reweighting terms. The learning rate is set at $5\times10^{-4}$, and a mini-batch size of 32 is used for gradient accumulation during model training. During the inference stage, the complete patch embedding sequence is passed directly into the MIL aggregator to obtain the predicted label (i.e., without TSS). Our code is publicly accessible via \url{https://github.com/LeiTong02/CLANet}.

\section{Results}
\label{sec:resutls}
In this section, we present the experimental results. First, we evaluate the quantitative performance of the proposed method on two data splitting tasks, comparing it to other related approaches. Next, we assess the effectiveness of different components within our proposed framework and investigate the influence of key parameters through ablation studies. Ultimately, we conduct a qualitative analysis as well as a failure case examination to provide a deeper insight into our proposed methodology.
\subsection{Quantitative Results}
To assess the efficiency of our proposed method in addressing batch effects during classification, we compare it against several recent MIL-based or domain adaptation methods.  Specifically, we consider the following methods: UDMIL (\cite{wang2020ud}) for optical coherence tomography volume classification, DSMIL(\cite{li2021dual}) for whole slide image classification, MCC(\cite{jin2020minimum}) for domain adaptation (DA), and DAMIL(\cite{hashimoto2020multi}), which incorporates DA with MIL for histopathology image classification. For a fair evaluation, the same backbone (ViT-S/8) is used in all methods with the input of CCS patches. Except for UDMIL, which we implemented ourselves, the other three methods use publicly available implementations. Additionally, we introduce ViT(w/ Image) and ViT(w/ CCS) as baselines to demonstrate the risk of batch effects for cell line classification. ViT(w/ Image) and ViT(w/ CCS) use raw cell images and CCS patches as input separately, along with augmentations (e.g., color jitters, vertical or horizontal flips). Both methods employ the ViT backbone with ImageNet weights. The three non-MIL methods, MCC, ViT(w/ Image), and ViT(w/ CCS), aggregate the predictions of images or patches through majority voting to obtain sequence or batch-level results. The comparison results are shown in Table \ref{tab:comparison_methods}.

Table \ref{tab:comparison_methods} demonstrates the substantial influence of batch effects on classification performance. While all methods achieve excellent results in the batch-stratified task, their performance noticeably decreases in the batch-separated task. For instance, Vision Transformer (ViT) with Image achieves 99.6\% batch-level accuracy when trained with data from all experimental batches. However, its performance significantly drops to 55.4\% batch-level accuracy in the batch-separated task. The implementation of our CCS patches as inputs has proven instrumental in enhancing classification performance. Even when merely used with ViT, CCS patches result in notable performance improvements compared to when raw images are used as inputs. DA-based methods, MCC and DAMIL, do not exhibit robust performance in addressing batch effects, potentially validating our hypothesis that batch effects cannot be fully mitigated by forceful alignment of source and target domains. Our proposed CLANet outperforms all, achieving the best overall performance with 89.1\% sequence-level and 90.7\% batch-level accuracy in the batch-separate task. The detailed prediction performance of CLANet is illustrated in the confusion matrices shown in Supplementary Fig. 2-4.

\subsection{Ablation Study}
\begin{figure*}[h]
\centering
    \includegraphics[width = 1\textwidth]{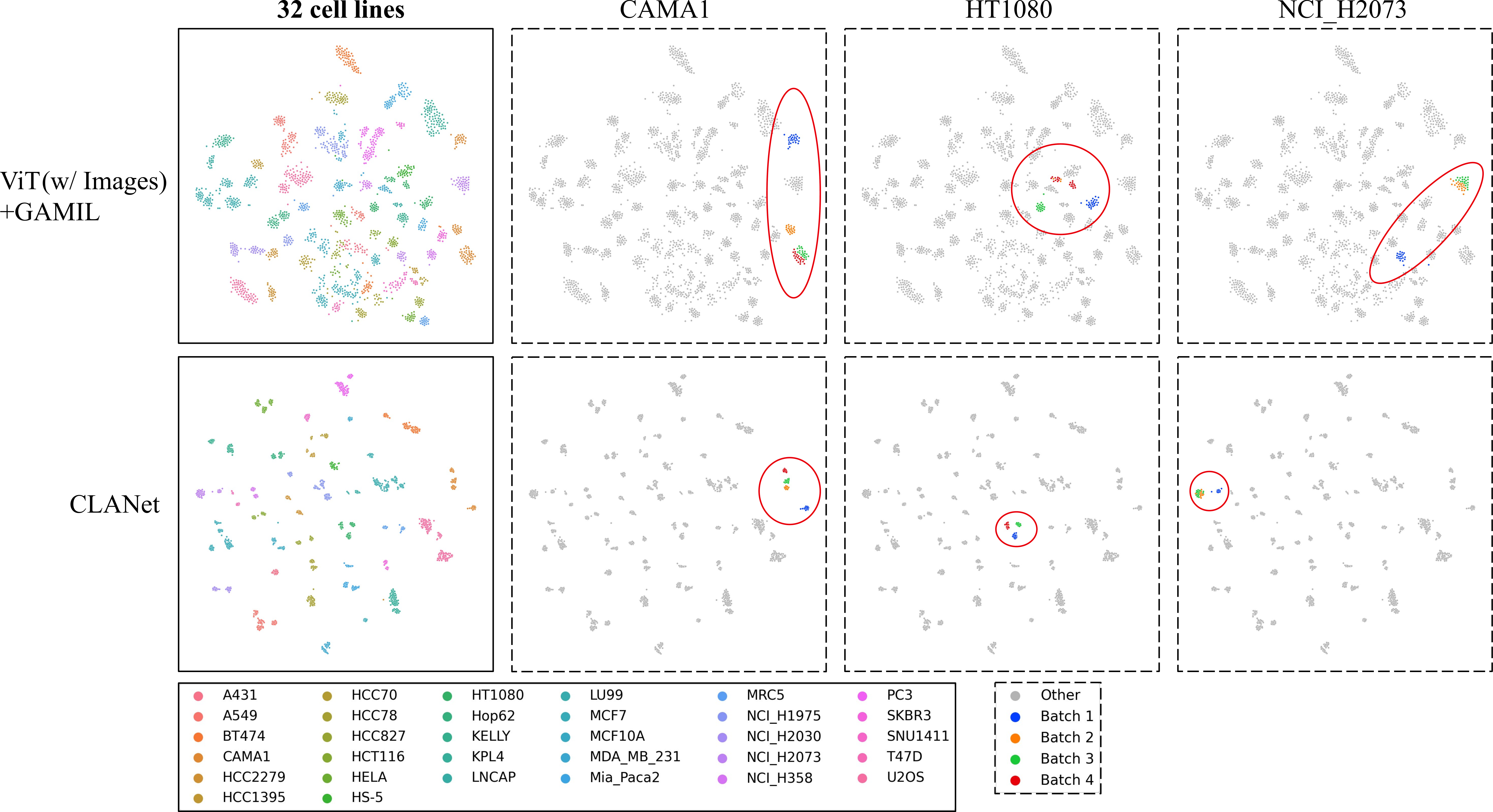}
    \caption{TSNE embeddings of the dataset. Each point represents an image sequence, with the corresponding feature vector captured from the MIL aggregator. The first column's figures illustrate the global distribution of 32 cell lines, while columns 2-4 showcase three example cell lines colored by their batch ID.}
    \label{fig:tsne_compar}
\end{figure*}
In this section, we conduct ablation studies to investigate the contributions of the key components of our framework on the batch-separated task. The following new notations are used to denote the different components and their combinations:
\begin{enumerate}
    \item {ViT(w/ Tiles): Represents the use of tiled patches as inputs instead of CCS patches.}
    \item{SSL: Indicates whether the model is fine-tuned using self-supervised learning (SSL).}
    \item{GAMIL: Denotes the incorporation of gated attention multiple instance learning (MIL).}
    \item{Max-Pooling/Avg-Pooling: Denotes the incorporation of max-pooling or average-pooling for feature aggregation.}
\end{enumerate}
The results are displayed in Table \ref{tab:ablation_study}. A comparison between ViT(w/ Tiles) and ViT(w/ CCS) demonstrates that using CCS patches as inputs significantly boosts classification accuracy. Interestingly, when comparing ViT(w/ CCS) with ViT(w/ CCS)+GAMIL, it becomes apparent that aggregating the results of instance predictions outperforms the utilization of GAMIL for aggregating instance features. Furthermore, leveraging self-supervised learning (SSL) for model fine-tuning aids in mitigating sample bias across various types of inputs, thereby further improving classification performance. Specifically, the integration of SSL with CCS patches leads to a substantial increase in sequence-level accuracy to 86.4\%(+16.2\%). Further integration with GAMIL results in enhanced accuracy and F1-score, suggesting that the patch features derived from SSL are representative of the target distribution and enable GAMIL to contribute effectively to overall performance.  Incorporating the TSS module provides an additional improvement of 2.1\%, 3.9\%, 2.8\%, and 2.8\% on seq-level and batch-level accuracy as well as F1-score. 

\begin{table}
\centering
\scriptsize
 \caption{Results of varying parameters on the batch-separated task. Default parameter settings are as follows: patch size (112$\times$112), patch number $K=10$, reweighting coefficients $(\alpha_{1}=1,\alpha_{2}=1)$. In each experiment, only one parameter is modified while the others remain consistent. It is important to note that when $\alpha_{1}=0$, the non-sampled embedding sequence will be unweighted to avoid assigning it a zero weight.}
\begin{tabular}{{lcccc}}
  \toprule
  \multirow{2}*{\shortstack[l]{Parameters}}&\multicolumn{2}{c}{Accuracy.\%}&\multicolumn{2}{c}{F1-score.\%}\\
  \cmidrule(lr){2-3}\cmidrule(lr){4-5}\ 
  & Sequence-level &Batch-level  & Sequence-level &Batch-level \\
  \midrule
  $51\times51$  & 77.8$\pm$4.6 & 77.7$\pm$4.0 & 67.9$\pm$6.5 & 71.9$\pm$5.6  \\
  
  $224\times224$  & 83.3$\pm$3.3 & 85.1$\pm$3.3 & 74.2$\pm$1.9 & 79.2$\pm$3.3  \\
   \midrule
  $K=5$   & 86.6$\pm$1.4 & 89.4$\pm$2.0 & 80.2$\pm$3.7 & 86.1$\pm$4.4  \\
  $K=20$  & 84.9$\pm$5.2 & 86.0$\pm$3.5 & 78.5$\pm$7.5 & 83.8$\pm$4.3  \\
  \midrule
  $\alpha_{1}=0,\alpha_{2}=0$   & 87.2$\pm$1.2 & 88.7$\pm$3.5 & 81.0$\pm$3.3 & 85.5$\pm$4.9  \\
  $\alpha_{1}=1,\alpha_{2}=0$   & 85.8$\pm$3.1 & 86.0$\pm$4.6 & 79.8$\pm$5.7 & 82.4$\pm$6.6  \\
  $\alpha_{1}=0,\alpha_{2}=1$   & 83.4$\pm$5.3 & 84.2$\pm$5.6 & 76.3$\pm$4.7 & 79.2$\pm$5.1  \\
  \midrule
  Default & \textbf{89.1$\pm$2.2} & \textbf{90.7$\pm$2.0} & \textbf{83.7$\pm$4.2} & \textbf{87.0$\pm$4.8}  \\
  \bottomrule
\end{tabular}
\label{tab:parameter_selection}
\end{table}

In Table \ref{tab:parameter_selection}, we assess the key parameter selection of our framework. Patch size significantly influences classification performance. Smaller patch sizes (51$\times$51) may not capture sufficient information from the cell patches, while larger patch sizes (224$\times$224) may include cells with more irregular shapes and a higher proportion of background pixels. Setting $K=10$ and enabling both reweighting terms allows the method to achieve optimal performance.
\begin{figure}[!ht]
\centering
    \includegraphics[width=\columnwidth]{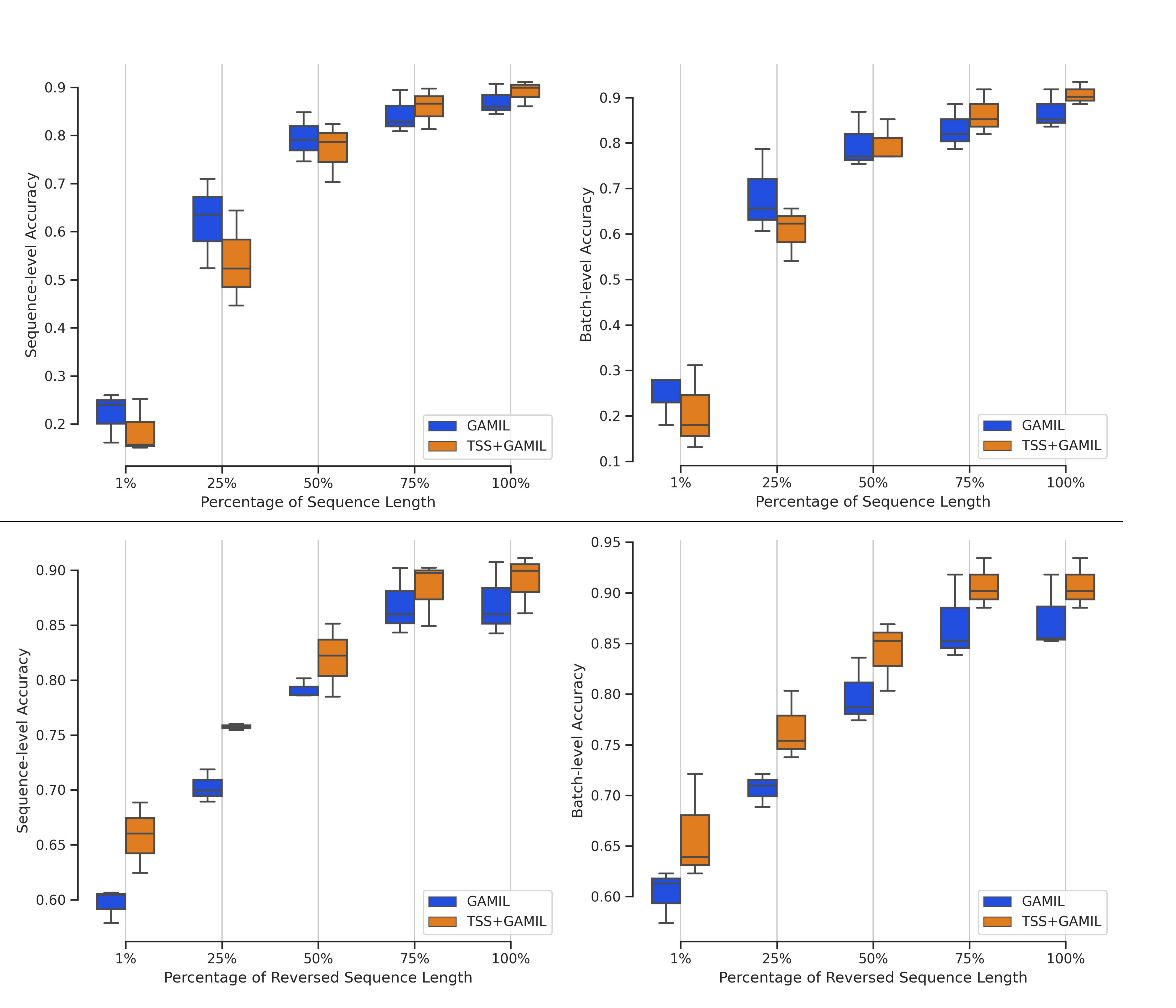}
    \caption{Classification performance against the percentage of sequence length utilized in the test set. The first row represents the sequences arranged in their natural order (i.e., from day 0 to day 12), while the second row illustrates the results for sequences arranged in reverse order.}
    \label{fig:box_plots}
\end{figure}
\subsection{Qualitative Analysis}
To validate the qualitative performance of our proposed method, we present the TSNE embeddings in comparison with the results obtained by ViT(w/ Images)+GAMIL, as shown in Figure \ref{fig:tsne_compar}. It is evident that CLANet yields a more compact distribution of samples, whereas ViT(w/ Images)+GAMIL generates a relatively sparse one. By highlighting three examples of cell lines along with their batch IDs, we observe that CLANet effectively maps samples of the same class into a cohesive region, thereby enhancing inter-class distance. Our method also maintains the intra-class distance across different batches, which suggests that our method is capable of preserving the discriminative information within each batch while also enhancing the separability between different cell lines.

Given the variability in image sequence lengths in our dataset, we explore the impact of sequence length on classification performance during testing. The percentage of sequence length utilized in the test set range in (1\%,25\%,50\%,75\%,100\%), as shown in Fig. \ref{fig:box_plots}. To ensure a comprehensive analysis, we compare the performance of our proposed methods against pure GAMIL, employing the same SSL features. Notably, as the percentage of sequence length utilized in testing increased, we consistently observed an improvement in classification performance. Moreover, sequences arranged in their natural order exhibited lower accuracy than those arranged in reverse order when a small percentage of sequence was used. This could be attributed to cell lines at early time points not having reached a stable state or displaying their characteristic morphology. In contrast, cells at later time points are more representative. Interestingly, GAMIL outperform our method when sequence lengths ranged from 1\% to 50\% in natural order, implying that GAMIL tends to focus more on samples at early incubation times. Nevertheless, our proposed TSS module can guide the GAMIL aggregator to gather more information from samples at later time points where cells display their true morphology, thus improving classification performance. This is validated by the results for sequences arranged in reverse order in Fig. \ref{fig:box_plots}. 
\begin{figure*}[h]
\centering
    \includegraphics[width = 1\textwidth]{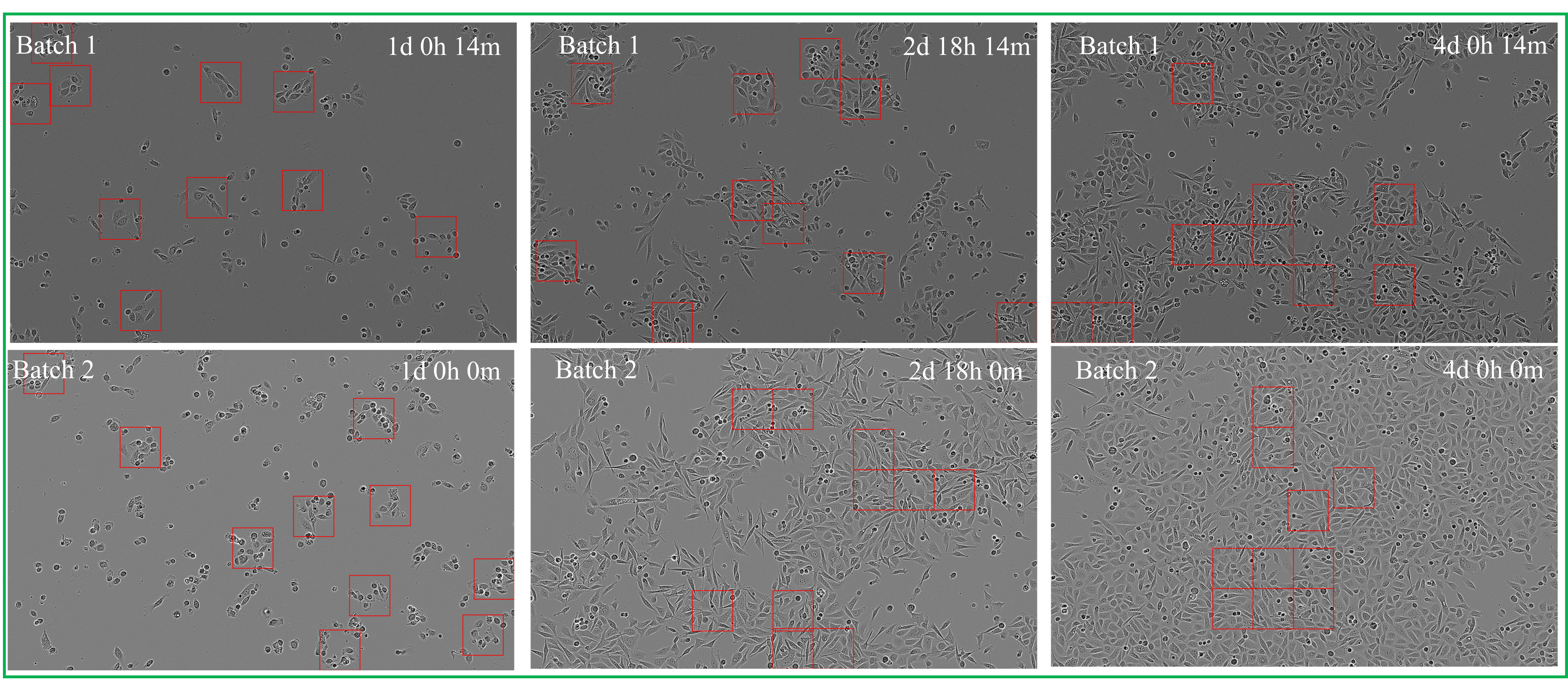}
    \caption{Visualization of the successful cases from the PC3 cell line. Red bounding boxes highlight the corresponding cell patches selected via the Cell Cluster-Level Selection (CCS) method.}
    \label{fig:successcase}
\end{figure*}
\begin{figure*}[!ht]
\centering
    \includegraphics[width = 1\textwidth]{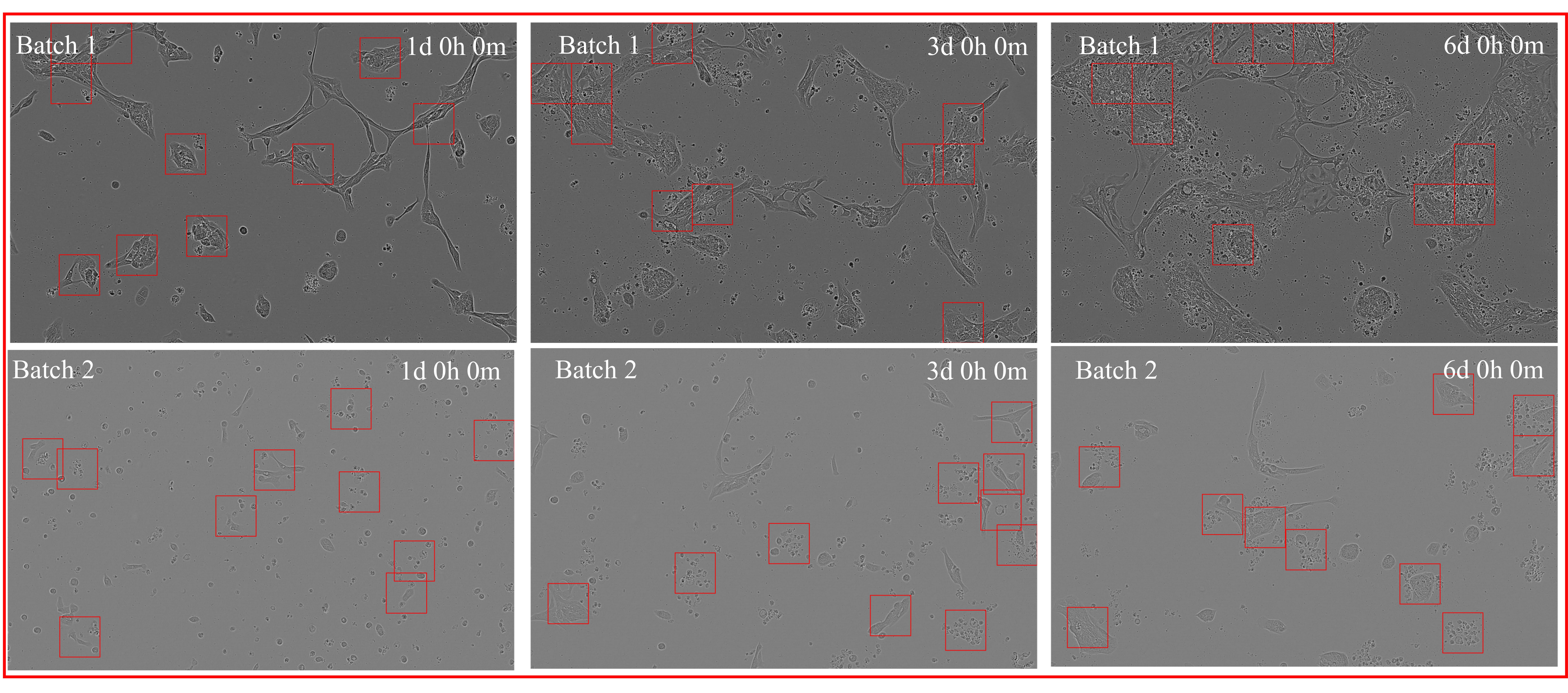}
    \caption{Visualization of the failure cases from the HCC70 cell line. Red bounding boxes highlight the corresponding cell patches selected via the Cell Cluster-Level Selection (CCS) method.}
    \label{fig:failcase}
\end{figure*}
\subsection{Analysis of Successful and Failed Cases}
To further understand the performance of our proposed method, we conduct an analysis of randomly selected successful and failed predictions, as depicted in Fig. \ref{fig:successcase}-\ref{fig:failcase}. Fig. \ref{fig:successcase} shows that despite variations in image quality and cell density between two batches of data, our method correctly classifies them regardless of which batch is used for training or testing (e.g., batch2 for training, batch 1 for testing). This implies that our proposed CCS method effectively captures representative patches across different batches, and the SSL strategy successfully addresses variations in image quality. Conversely, the failed predictions, as shown in Fig. \ref{fig:failcase}, provide insight into potential areas of enhancement. The batch 2 image reveals unhealthy cell growth, with numerous cells displaying almost circular morphologies indicative of cell death. In contrast, cells in batch 1 display their true, elongated morphologies. These failed cases imply that enhancing our patch selection method to prioritize healthy cells could improve classification performance. This improvement could potentially be achieved by incorporating anomaly detection techniques into our methodology.

\section{Conclusion}
\label{sec:conclusion}
In this study, we present a novel framework specifically tailored for cross-batch cell line identification using brightfield images. We postulate three distinct forms of batch effects and propose targeted solutions for each, marking a pioneering effort in addressing cross-batch cell line identification based on brightfield images.

Our framework incorporates a cell-cluster level selection strategy to capture representative cell patches, accounting for the impact of cell density on cell images. To mitigate the risk of batch-effect-induced biased predictions and to generate robust patch embeddings, we introduce a self-supervised learning strategy. Further, we adopt multiple instance learning to fuse patch embeddings for cell line identification and propose a novel time-series segment sampling module designed to address the bias introduced by varying incubation times among cell batches.

To support our research, we have amassed a large-scale dataset from Astrazeneca Global Cell Bank. Through comprehensive analysis and ablation studies, we demonstrate the effectiveness of our proposed framework in handling batch-separated tasks. Each component of our framework contributes significantly to the overall performance, showcasing its effectiveness in tackling the challenges inherent in cross-batch cell line identification based on brightfield images.

\section*{CRediT authorship contribution statement}
\textbf{Lei Tong}: Conceptualization, Methodology, Data curation, Software, Validation, Formal analysis, Investigation, Writing – original
draft, Writing – review $\&$ editing, Visualization. \textbf{Adam Corrigan}: Conceptualization, Methodology, Writing – review $\&$ editing. \textbf{Navin Rathna Kumar}: Data curation, Investigation, Writing – review $\&$ editing. \textbf{Kerry Hallbrook}: Data curation, Investigation, Writing – review $\&$ editing. \textbf{Jonathan Orme}: Writing – review $\&$ editing, Supervision, Funding acquisition. \textbf{Yinhai Wang}: Conceptualization, Writing – review $\&$ editing, Supervision, Funding acquisition, Project administration. \textbf{Huiyu Zhou}: Conceptualization, Writing – review $\&$ editing, Supervision, Funding acquisition.

\section*{Declaration of competing interest}
The authors declare that they have no known competing financial interests or personal relationships that could have appeared to influence the work reported in this paper.

\section*{Acknowledgement}
Authors Lei Tong is sponsored by University of Leicester GTA studentship (GTA 2020), China Scholarship Council and AstraZeneca – University Leicester collaboration agreement (CR-019972). Authors (Adam Corrigan, Navin Rathna Kumar, Kerry Hallbrook, Jonathan Orme, Yinhai Wang) are employees of AstraZeneca. AstraZeneca provided the funding for this research and provided support in the form of salaries for all authors, but did not have any additional role in the study design, data collection and analysis, decision to publish, or preparation of the manuscript.

\bibliographystyle{model2-names.bst}\biboptions{authoryear}
\bibliography{refs}

\begin{thebibliography}{37}
\expandafter\ifx\csname natexlab\endcsname\relax\def\natexlab#1{#1}\fi
\providecommand{\url}[1]{\texttt{#1}}
\providecommand{\href}[2]{#2}
\providecommand{\path}[1]{#1}
\providecommand{\DOIprefix}{doi:}
\providecommand{\ArXivprefix}{arXiv:}
\providecommand{\URLprefix}{URL: }
\providecommand{\Pubmedprefix}{pmid:}
\providecommand{\doi}[1]{\href{http://dx.doi.org/#1}{\path{#1}}}
\providecommand{\Pubmed}[1]{\href{pmid:#1}{\path{#1}}}
\providecommand{\bibinfo}[2]{#2}
\ifx\xfnm\relax \def\xfnm[#1]{\unskip,\space#1}\fi
\bibitem[{Ando et~al.(2017)Ando, McLean and Berndl}]{ando2017improving}
\bibinfo{author}{Ando, D.M.}, \bibinfo{author}{McLean, C.Y.},
  \bibinfo{author}{Berndl, M.}, \bibinfo{year}{2017}.
\newblock \bibinfo{title}{Improving phenotypic measurements in high-content
  imaging screens}.
\newblock \bibinfo{journal}{BioRxiv} , \bibinfo{pages}{161422}.
\bibitem[{Belashov et~al.(2021)Belashov, Zhikhoreva, Belyaeva, Salova,
  Kornilova, Semenova and Vasyutinskii}]{belashov2021machine}
\bibinfo{author}{Belashov, A.V.}, \bibinfo{author}{Zhikhoreva, A.A.},
  \bibinfo{author}{Belyaeva, T.N.}, \bibinfo{author}{Salova, A.V.},
  \bibinfo{author}{Kornilova, E.S.}, \bibinfo{author}{Semenova, I.V.},
  \bibinfo{author}{Vasyutinskii, O.S.}, \bibinfo{year}{2021}.
\newblock \bibinfo{title}{Machine learning assisted classification of cell
  lines and cell states on quantitative phase images}.
\newblock \bibinfo{journal}{Cells} \bibinfo{volume}{10}, \bibinfo{pages}{2587}.
\bibitem[{Boonstra et~al.(2010)Boonstra, Van~Marion, Beer, Lin, Chaves,
  Ribeiro, Pereira, Roque, Darnton, Altorki et~al.}]{boonstra2010verification}
\bibinfo{author}{Boonstra, J.J.}, \bibinfo{author}{Van~Marion, R.},
  \bibinfo{author}{Beer, D.G.}, \bibinfo{author}{Lin, L.},
  \bibinfo{author}{Chaves, P.}, \bibinfo{author}{Ribeiro, C.},
  \bibinfo{author}{Pereira, A.D.}, \bibinfo{author}{Roque, L.},
  \bibinfo{author}{Darnton, S.J.}, \bibinfo{author}{Altorki, N.K.}, et~al.,
  \bibinfo{year}{2010}.
\newblock \bibinfo{title}{Verification and unmasking of widely used human
  esophageal adenocarcinoma cell lines}.
\newblock \bibinfo{journal}{Journal of the National Cancer Institute}
  \bibinfo{volume}{102}, \bibinfo{pages}{271--274}.
\bibitem[{Buggenthin et~al.(2013)Buggenthin, Marr, Schwarzfischer, Hoppe,
  Hilsenbeck, Schroeder and Theis}]{buggenthin2013automatic}
\bibinfo{author}{Buggenthin, F.}, \bibinfo{author}{Marr, C.},
  \bibinfo{author}{Schwarzfischer, M.}, \bibinfo{author}{Hoppe, P.S.},
  \bibinfo{author}{Hilsenbeck, O.}, \bibinfo{author}{Schroeder, T.},
  \bibinfo{author}{Theis, F.J.}, \bibinfo{year}{2013}.
\newblock \bibinfo{title}{An automatic method for robust and fast cell
  detection in bright field images from high-throughput microscopy}.
\newblock \bibinfo{journal}{BMC bioinformatics} \bibinfo{volume}{14},
  \bibinfo{pages}{1--12}.
\bibitem[{Caicedo et~al.(2017)Caicedo, Cooper, Heigwer, Warchal, Qiu, Molnar,
  Vasilevich, Barry, Bansal, Kraus et~al.}]{caicedo2017data}
\bibinfo{author}{Caicedo, J.C.}, \bibinfo{author}{Cooper, S.},
  \bibinfo{author}{Heigwer, F.}, \bibinfo{author}{Warchal, S.},
  \bibinfo{author}{Qiu, P.}, \bibinfo{author}{Molnar, C.},
  \bibinfo{author}{Vasilevich, A.S.}, \bibinfo{author}{Barry, J.D.},
  \bibinfo{author}{Bansal, H.S.}, \bibinfo{author}{Kraus, O.}, et~al.,
  \bibinfo{year}{2017}.
\newblock \bibinfo{title}{Data-analysis strategies for image-based cell
  profiling}.
\newblock \bibinfo{journal}{Nature methods} \bibinfo{volume}{14},
  \bibinfo{pages}{849--863}.
\bibitem[{Caicedo et~al.(2016)Caicedo, Singh and
  Carpenter}]{caicedo2016applications}
\bibinfo{author}{Caicedo, J.C.}, \bibinfo{author}{Singh, S.},
  \bibinfo{author}{Carpenter, A.E.}, \bibinfo{year}{2016}.
\newblock \bibinfo{title}{Applications in image-based profiling of
  perturbations}.
\newblock \bibinfo{journal}{Current opinion in biotechnology}
  \bibinfo{volume}{39}, \bibinfo{pages}{134--142}.
\bibitem[{Caron et~al.(2020)Caron, Misra, Mairal, Goyal, Bojanowski and
  Joulin}]{caron2020unsupervised}
\bibinfo{author}{Caron, M.}, \bibinfo{author}{Misra, I.},
  \bibinfo{author}{Mairal, J.}, \bibinfo{author}{Goyal, P.},
  \bibinfo{author}{Bojanowski, P.}, \bibinfo{author}{Joulin, A.},
  \bibinfo{year}{2020}.
\newblock \bibinfo{title}{Unsupervised learning of visual features by
  contrasting cluster assignments}.
\newblock \bibinfo{journal}{Advances in neural information processing systems}
  \bibinfo{volume}{33}, \bibinfo{pages}{9912--9924}.
\bibitem[{Caron et~al.(2021)Caron, Touvron, Misra, J{\'e}gou, Mairal,
  Bojanowski and Joulin}]{caron2021emerging}
\bibinfo{author}{Caron, M.}, \bibinfo{author}{Touvron, H.},
  \bibinfo{author}{Misra, I.}, \bibinfo{author}{J{\'e}gou, H.},
  \bibinfo{author}{Mairal, J.}, \bibinfo{author}{Bojanowski, P.},
  \bibinfo{author}{Joulin, A.}, \bibinfo{year}{2021}.
\newblock \bibinfo{title}{Emerging properties in self-supervised vision
  transformers}, in: \bibinfo{booktitle}{Proceedings of the IEEE/CVF
  international conference on computer vision}, pp.
  \bibinfo{pages}{9650--9660}.
\bibitem[{Chandrasekaran et~al.(2021)Chandrasekaran, Ceulemans, Boyd and
  Carpenter}]{chandrasekaran2021image}
\bibinfo{author}{Chandrasekaran, S.N.}, \bibinfo{author}{Ceulemans, H.},
  \bibinfo{author}{Boyd, J.D.}, \bibinfo{author}{Carpenter, A.E.},
  \bibinfo{year}{2021}.
\newblock \bibinfo{title}{Image-based profiling for drug discovery: due for a
  machine-learning upgrade?}
\newblock \bibinfo{journal}{Nature Reviews Drug Discovery}
  \bibinfo{volume}{20}, \bibinfo{pages}{145--159}.
\bibitem[{Chikontwe et~al.(2021)Chikontwe, Luna, Kang, Hong, Ahn and
  Park}]{chikontwe2021dual}
\bibinfo{author}{Chikontwe, P.}, \bibinfo{author}{Luna, M.},
  \bibinfo{author}{Kang, M.}, \bibinfo{author}{Hong, K.S.},
  \bibinfo{author}{Ahn, J.H.}, \bibinfo{author}{Park, S.H.},
  \bibinfo{year}{2021}.
\newblock \bibinfo{title}{Dual attention multiple instance learning with
  unsupervised complementary loss for covid-19 screening}.
\newblock \bibinfo{journal}{Medical Image Analysis} \bibinfo{volume}{72},
  \bibinfo{pages}{102105}.
\bibitem[{Cross-Zamirski et~al.(2022)Cross-Zamirski, Williams, Mouchet,
  Sch{\"o}nlieb, Turkki and Wang}]{cross2022self}
\bibinfo{author}{Cross-Zamirski, J.O.}, \bibinfo{author}{Williams, G.},
  \bibinfo{author}{Mouchet, E.}, \bibinfo{author}{Sch{\"o}nlieb, C.B.},
  \bibinfo{author}{Turkki, R.}, \bibinfo{author}{Wang, Y.},
  \bibinfo{year}{2022}.
\newblock \bibinfo{title}{Self-supervised learning of phenotypic
  representations from cell images with weak labels}.
\newblock \bibinfo{journal}{arXiv preprint arXiv:2209.07819} .
\bibitem[{Freedman et~al.(2015)Freedman, Gibson, Ethier, Soule, Neve and
  Reid}]{freedman2015reproducibility}
\bibinfo{author}{Freedman, L.P.}, \bibinfo{author}{Gibson, M.C.},
  \bibinfo{author}{Ethier, S.P.}, \bibinfo{author}{Soule, H.R.},
  \bibinfo{author}{Neve, R.M.}, \bibinfo{author}{Reid, Y.A.},
  \bibinfo{year}{2015}.
\newblock \bibinfo{title}{Reproducibility: changing the policies and culture of
  cell line authentication}.
\newblock \bibinfo{journal}{Nature methods} \bibinfo{volume}{12},
  \bibinfo{pages}{493--497}.
\bibitem[{Hashimoto et~al.(2020)Hashimoto, Fukushima, Koga, Takagi, Ko, Kohno,
  Nakaguro, Nakamura, Hontani and Takeuchi}]{hashimoto2020multi}
\bibinfo{author}{Hashimoto, N.}, \bibinfo{author}{Fukushima, D.},
  \bibinfo{author}{Koga, R.}, \bibinfo{author}{Takagi, Y.},
  \bibinfo{author}{Ko, K.}, \bibinfo{author}{Kohno, K.},
  \bibinfo{author}{Nakaguro, M.}, \bibinfo{author}{Nakamura, S.},
  \bibinfo{author}{Hontani, H.}, \bibinfo{author}{Takeuchi, I.},
  \bibinfo{year}{2020}.
\newblock \bibinfo{title}{Multi-scale domain-adversarial multiple-instance cnn
  for cancer subtype classification with unannotated histopathological images},
  in: \bibinfo{booktitle}{Proceedings of the IEEE/CVF conference on computer
  vision and pattern recognition}, pp. \bibinfo{pages}{3852--3861}.
\bibitem[{Ilse et~al.(2018)Ilse, Tomczak and Welling}]{ilse2018attention}
\bibinfo{author}{Ilse, M.}, \bibinfo{author}{Tomczak, J.},
  \bibinfo{author}{Welling, M.}, \bibinfo{year}{2018}.
\newblock \bibinfo{title}{Attention-based deep multiple instance learning}, in:
  \bibinfo{booktitle}{International conference on machine learning},
  \bibinfo{organization}{PMLR}. pp. \bibinfo{pages}{2127--2136}.
\bibitem[{Ioannidis(2005)}]{ioannidis2005most}
\bibinfo{author}{Ioannidis, J.P.}, \bibinfo{year}{2005}.
\newblock \bibinfo{title}{Why most published research findings are false}.
\newblock \bibinfo{journal}{PLoS medicine} \bibinfo{volume}{2},
  \bibinfo{pages}{e124}.
\bibitem[{Janssens et~al.(2013)Janssens, Antanas, Derde, Vanhorebeek, Van~den
  Berghe and Grandas}]{janssens2013charisma}
\bibinfo{author}{Janssens, T.}, \bibinfo{author}{Antanas, L.},
  \bibinfo{author}{Derde, S.}, \bibinfo{author}{Vanhorebeek, I.},
  \bibinfo{author}{Van~den Berghe, G.}, \bibinfo{author}{Grandas, F.G.},
  \bibinfo{year}{2013}.
\newblock \bibinfo{title}{Charisma: An integrated approach to automatic
  h\&e-stained skeletal muscle cell segmentation using supervised learning and
  novel robust clump splitting}.
\newblock \bibinfo{journal}{Medical image analysis} \bibinfo{volume}{17},
  \bibinfo{pages}{1206--1219}.
\bibitem[{Jin et~al.(2020)Jin, Wang, Long and Wang}]{jin2020minimum}
\bibinfo{author}{Jin, Y.}, \bibinfo{author}{Wang, X.}, \bibinfo{author}{Long,
  M.}, \bibinfo{author}{Wang, J.}, \bibinfo{year}{2020}.
\newblock \bibinfo{title}{Minimum class confusion for versatile domain
  adaptation}, in: \bibinfo{booktitle}{Computer Vision--ECCV 2020: 16th
  European Conference, Glasgow, UK, August 23--28, 2020, Proceedings, Part XXI
  16}, \bibinfo{organization}{Springer}. pp. \bibinfo{pages}{464--480}.
\bibitem[{Kirillov et~al.(2023)Kirillov, Mintun, Ravi, Mao, Rolland, Gustafson,
  Xiao, Whitehead, Berg, Lo et~al.}]{kirillov2023segment}
\bibinfo{author}{Kirillov, A.}, \bibinfo{author}{Mintun, E.},
  \bibinfo{author}{Ravi, N.}, \bibinfo{author}{Mao, H.},
  \bibinfo{author}{Rolland, C.}, \bibinfo{author}{Gustafson, L.},
  \bibinfo{author}{Xiao, T.}, \bibinfo{author}{Whitehead, S.},
  \bibinfo{author}{Berg, A.C.}, \bibinfo{author}{Lo, W.Y.}, et~al.,
  \bibinfo{year}{2023}.
\newblock \bibinfo{title}{Segment anything}.
\newblock \bibinfo{journal}{arXiv preprint arXiv:2304.02643} .
\bibitem[{Kothari et~al.(2013)Kothari, Phan, Stokes, Osunkoya, Young and
  Wang}]{kothari2013removing}
\bibinfo{author}{Kothari, S.}, \bibinfo{author}{Phan, J.H.},
  \bibinfo{author}{Stokes, T.H.}, \bibinfo{author}{Osunkoya, A.O.},
  \bibinfo{author}{Young, A.N.}, \bibinfo{author}{Wang, M.D.},
  \bibinfo{year}{2013}.
\newblock \bibinfo{title}{Removing batch effects from histopathological images
  for enhanced cancer diagnosis}.
\newblock \bibinfo{journal}{IEEE journal of biomedical and health informatics}
  \bibinfo{volume}{18}, \bibinfo{pages}{765--772}.
\bibitem[{Li et~al.(2021)Li, Li and Eliceiri}]{li2021dual}
\bibinfo{author}{Li, B.}, \bibinfo{author}{Li, Y.}, \bibinfo{author}{Eliceiri,
  K.W.}, \bibinfo{year}{2021}.
\newblock \bibinfo{title}{Dual-stream multiple instance learning network for
  whole slide image classification with self-supervised contrastive learning},
  in: \bibinfo{booktitle}{Proceedings of the IEEE/CVF conference on computer
  vision and pattern recognition}, pp. \bibinfo{pages}{14318--14328}.
\bibitem[{Li et~al.(2020)Li, Wang, Lyu, Pan, Zhang, Stambolian, Susztak,
  Reilly, Hu and Li}]{li2020deep}
\bibinfo{author}{Li, X.}, \bibinfo{author}{Wang, K.}, \bibinfo{author}{Lyu,
  Y.}, \bibinfo{author}{Pan, H.}, \bibinfo{author}{Zhang, J.},
  \bibinfo{author}{Stambolian, D.}, \bibinfo{author}{Susztak, K.},
  \bibinfo{author}{Reilly, M.P.}, \bibinfo{author}{Hu, G.},
  \bibinfo{author}{Li, M.}, \bibinfo{year}{2020}.
\newblock \bibinfo{title}{Deep learning enables accurate clustering with batch
  effect removal in single-cell rna-seq analysis}.
\newblock \bibinfo{journal}{Nature communications} \bibinfo{volume}{11},
  \bibinfo{pages}{2338}.
\bibitem[{Masters et~al.(2001)Masters, Thomson, Daly-Burns, Reid, Dirks,
  Packer, Toji, Ohno, Tanabe, Arlett et~al.}]{masters2001short}
\bibinfo{author}{Masters, J.R.}, \bibinfo{author}{Thomson, J.A.},
  \bibinfo{author}{Daly-Burns, B.}, \bibinfo{author}{Reid, Y.A.},
  \bibinfo{author}{Dirks, W.G.}, \bibinfo{author}{Packer, P.},
  \bibinfo{author}{Toji, L.H.}, \bibinfo{author}{Ohno, T.},
  \bibinfo{author}{Tanabe, H.}, \bibinfo{author}{Arlett, C.F.}, et~al.,
  \bibinfo{year}{2001}.
\newblock \bibinfo{title}{Short tandem repeat profiling provides an
  international reference standard for human cell lines}.
\newblock \bibinfo{journal}{Proceedings of the National Academy of Sciences}
  \bibinfo{volume}{98}, \bibinfo{pages}{8012--8017}.
\bibitem[{Mzurikwao et~al.(2020)Mzurikwao, Khan, Samuel, Cinatl, Wass,
  Michaelis, Marcelli and Ang}]{mzurikwao2020towards}
\bibinfo{author}{Mzurikwao, D.}, \bibinfo{author}{Khan, M.U.},
  \bibinfo{author}{Samuel, O.W.}, \bibinfo{author}{Cinatl, J.},
  \bibinfo{author}{Wass, M.}, \bibinfo{author}{Michaelis, M.},
  \bibinfo{author}{Marcelli, G.}, \bibinfo{author}{Ang, C.S.},
  \bibinfo{year}{2020}.
\newblock \bibinfo{title}{Towards image-based cancer cell lines authentication
  using deep neural networks}.
\newblock \bibinfo{journal}{Scientific Reports} \bibinfo{volume}{10},
  \bibinfo{pages}{1--15}.
\bibitem[{Ojala et~al.(2002)Ojala, Pietikainen and
  Maenpaa}]{ojala2002multiresolution}
\bibinfo{author}{Ojala, T.}, \bibinfo{author}{Pietikainen, M.},
  \bibinfo{author}{Maenpaa, T.}, \bibinfo{year}{2002}.
\newblock \bibinfo{title}{Multiresolution gray-scale and rotation invariant
  texture classification with local binary patterns}.
\newblock \bibinfo{journal}{IEEE Transactions on pattern analysis and machine
  intelligence} \bibinfo{volume}{24}, \bibinfo{pages}{971--987}.
\bibitem[{Pachitariu and Stringer(2022)}]{pachitariu2022cellpose}
\bibinfo{author}{Pachitariu, M.}, \bibinfo{author}{Stringer, C.},
  \bibinfo{year}{2022}.
\newblock \bibinfo{title}{Cellpose 2.0: how to train your own model}.
\newblock \bibinfo{journal}{Nature Methods} , \bibinfo{pages}{1--8}.
\bibitem[{Parson et~al.(2005)Parson, Kirchebner, M{\"u}hlmann, Renner, Kofler,
  Schmidt and Kofler}]{parson2005cancer}
\bibinfo{author}{Parson, W.}, \bibinfo{author}{Kirchebner, R.},
  \bibinfo{author}{M{\"u}hlmann, R.}, \bibinfo{author}{Renner, K.},
  \bibinfo{author}{Kofler, A.}, \bibinfo{author}{Schmidt, S.},
  \bibinfo{author}{Kofler, R.}, \bibinfo{year}{2005}.
\newblock \bibinfo{title}{Cancer cell line identification by short tandem
  repeat profiling: power and limitations}.
\newblock \bibinfo{journal}{The FASEB journal} \bibinfo{volume}{19},
  \bibinfo{pages}{1--18}.
\bibitem[{Reid et~al.(2013)Reid, Storts, Riss and
  Minor}]{reid2013authentication}
\bibinfo{author}{Reid, Y.}, \bibinfo{author}{Storts, D.},
  \bibinfo{author}{Riss, T.}, \bibinfo{author}{Minor, L.},
  \bibinfo{year}{2013}.
\newblock \bibinfo{title}{Authentication of human cell lines by str dna
  profiling analysis}.
\newblock \bibinfo{journal}{Assay guidance manual [Internet]} .
\bibitem[{Su et~al.(2022)Su, Tavolara, Carreno-Galeano, Lee, Gurcan and
  Niazi}]{su2022attention2majority}
\bibinfo{author}{Su, Z.}, \bibinfo{author}{Tavolara, T.E.},
  \bibinfo{author}{Carreno-Galeano, G.}, \bibinfo{author}{Lee, S.J.},
  \bibinfo{author}{Gurcan, M.N.}, \bibinfo{author}{Niazi, M.},
  \bibinfo{year}{2022}.
\newblock \bibinfo{title}{Attention2majority: Weak multiple instance learning
  for regenerative kidney grading on whole slide images}.
\newblock \bibinfo{journal}{Medical Image Analysis} \bibinfo{volume}{79},
  \bibinfo{pages}{102462}.
\bibitem[{Sypetkowski et~al.(2023)Sypetkowski, Rezanejad, Saberian, Kraus,
  Urbanik, Taylor, Mabey, Victors, Yosinski, Sereshkeh
  et~al.}]{sypetkowski2023rxrx1}
\bibinfo{author}{Sypetkowski, M.}, \bibinfo{author}{Rezanejad, M.},
  \bibinfo{author}{Saberian, S.}, \bibinfo{author}{Kraus, O.},
  \bibinfo{author}{Urbanik, J.}, \bibinfo{author}{Taylor, J.},
  \bibinfo{author}{Mabey, B.}, \bibinfo{author}{Victors, M.},
  \bibinfo{author}{Yosinski, J.}, \bibinfo{author}{Sereshkeh, A.R.}, et~al.,
  \bibinfo{year}{2023}.
\newblock \bibinfo{title}{Rxrx1: A dataset for evaluating experimental batch
  correction methods}.
\newblock \bibinfo{journal}{arXiv preprint arXiv:2301.05768} .
\bibitem[{Tong et~al.(2022)Tong, Corrigan, Kumar, Hallbrook, Orme, Wang and
  Zhou}]{tong2022automated}
\bibinfo{author}{Tong, L.}, \bibinfo{author}{Corrigan, A.},
  \bibinfo{author}{Kumar, N.R.}, \bibinfo{author}{Hallbrook, K.},
  \bibinfo{author}{Orme, J.}, \bibinfo{author}{Wang, Y.},
  \bibinfo{author}{Zhou, H.}, \bibinfo{year}{2022}.
\newblock \bibinfo{title}{An automated cell line authentication method for
  astrazeneca global cell bank using deep neural networks on brightfield
  images}.
\newblock \bibinfo{journal}{Scientific reports} \bibinfo{volume}{12},
  \bibinfo{pages}{1--11}.
\bibitem[{Uijlings et~al.(2013)Uijlings, Van De~Sande, Gevers and
  Smeulders}]{uijlings2013selective}
\bibinfo{author}{Uijlings, J.R.}, \bibinfo{author}{Van De~Sande, K.E.},
  \bibinfo{author}{Gevers, T.}, \bibinfo{author}{Smeulders, A.W.},
  \bibinfo{year}{2013}.
\newblock \bibinfo{title}{Selective search for object recognition}.
\newblock \bibinfo{journal}{International journal of computer vision}
  \bibinfo{volume}{104}, \bibinfo{pages}{154--171}.
\bibitem[{Wang et~al.(2018)Wang, Xiong, Wang, Qiao, Lin, Tang and
  Van~Gool}]{wang2018temporal}
\bibinfo{author}{Wang, L.}, \bibinfo{author}{Xiong, Y.}, \bibinfo{author}{Wang,
  Z.}, \bibinfo{author}{Qiao, Y.}, \bibinfo{author}{Lin, D.},
  \bibinfo{author}{Tang, X.}, \bibinfo{author}{Van~Gool, L.},
  \bibinfo{year}{2018}.
\newblock \bibinfo{title}{Temporal segment networks for action recognition in
  videos}.
\newblock \bibinfo{journal}{IEEE transactions on pattern analysis and machine
  intelligence} \bibinfo{volume}{41}, \bibinfo{pages}{2740--2755}.
\bibitem[{Wang et~al.(2020a)Wang, Wang, Kang, Guo, Guo, Dongye, Zhu, Chen,
  Zhang, Long et~al.}]{wang2020artificial}
\bibinfo{author}{Wang, R.}, \bibinfo{author}{Wang, D.}, \bibinfo{author}{Kang,
  D.}, \bibinfo{author}{Guo, X.}, \bibinfo{author}{Guo, C.},
  \bibinfo{author}{Dongye, M.}, \bibinfo{author}{Zhu, Y.},
  \bibinfo{author}{Chen, C.}, \bibinfo{author}{Zhang, X.},
  \bibinfo{author}{Long, E.}, et~al., \bibinfo{year}{2020}a.
\newblock \bibinfo{title}{An artificial intelligent platform for live cell
  identification and the detection of cross-contamination}.
\newblock \bibinfo{journal}{Annals of Translational Medicine}
  \bibinfo{volume}{8}.
\bibitem[{Wang et~al.(2023)Wang, Tang, Chen, Cheung and Heng}]{wang2023deep}
\bibinfo{author}{Wang, X.}, \bibinfo{author}{Tang, F.}, \bibinfo{author}{Chen,
  H.}, \bibinfo{author}{Cheung, C.Y.}, \bibinfo{author}{Heng, P.A.},
  \bibinfo{year}{2023}.
\newblock \bibinfo{title}{Deep semi-supervised multiple instance learning with
  self-correction for dme classification from oct images}.
\newblock \bibinfo{journal}{Medical Image Analysis} \bibinfo{volume}{83},
  \bibinfo{pages}{102673}.
\bibitem[{Wang et~al.(2020b)Wang, Tang, Chen, Luo, Tang, Ran, Cheung and
  Heng}]{wang2020ud}
\bibinfo{author}{Wang, X.}, \bibinfo{author}{Tang, F.}, \bibinfo{author}{Chen,
  H.}, \bibinfo{author}{Luo, L.}, \bibinfo{author}{Tang, Z.},
  \bibinfo{author}{Ran, A.R.}, \bibinfo{author}{Cheung, C.Y.},
  \bibinfo{author}{Heng, P.A.}, \bibinfo{year}{2020}b.
\newblock \bibinfo{title}{Ud-mil: uncertainty-driven deep multiple instance
  learning for oct image classification}.
\newblock \bibinfo{journal}{IEEE journal of biomedical and health informatics}
  \bibinfo{volume}{24}, \bibinfo{pages}{3431--3442}.
\bibitem[{Wu et~al.(2005)Wu, Otoo and Shoshani}]{wu2005optimizing}
\bibinfo{author}{Wu, K.}, \bibinfo{author}{Otoo, E.},
  \bibinfo{author}{Shoshani, A.}, \bibinfo{year}{2005}.
\newblock \bibinfo{title}{Optimizing connected component labeling algorithms},
  in: \bibinfo{booktitle}{Medical Imaging 2005: Image Processing},
  \bibinfo{organization}{SPIE}. pp. \bibinfo{pages}{1965--1976}.
\bibitem[{Yao et~al.(2019)Yao, Rochman and Sun}]{yao2019cell}
\bibinfo{author}{Yao, K.}, \bibinfo{author}{Rochman, N.D.},
  \bibinfo{author}{Sun, S.X.}, \bibinfo{year}{2019}.
\newblock \bibinfo{title}{Cell type classification and unsupervised
  morphological phenotyping from low-resolution images using deep learning}.
\newblock \bibinfo{journal}{Scientific reports} \bibinfo{volume}{9},
  \bibinfo{pages}{1--13}.

\end{thebibliography}

\end{document}